\documentclass[wcp]{jmlr}
\pdfoutput=1
\jmlrvolume{230}
\jmlryear{2024}
\jmlrworkshop{Conformal and Probabilistic Prediction with Applications}
\jmlrproceedings{PMLR}{Proceedings of Machine Learning Research}

\usepackage{amsmath,amssymb,graphicx,url}
\usepackage{rotating}% for sideways figures and tables
\usepackage{longtable}% for long tables

\usepackage{booktabs}
\usepackage[load-configurations=version-1]{siunitx} % newer version
\usepackage{bbm}
\usepackage{multirow} 
\usepackage{subcaption}
\usepackage{float}

\theorembodyfont{\upshape}
\theoremheaderfont{\scshape}
\theorempostheader{:}
\theoremsep{\newline}

\title[Evidential Conformal Prediction]{Evidential Uncertainty Sets in Deep Classifiers Using Conformal Prediction}
\author{\Name{Hamed Karimi}\Email{hamed.karimi@torontomu.ca}\\
\addr{Department of Electrical, Computer, and Biomedical Engineering, Toronto Metropolitan University, Toronto, ON, Canada}\\
\Name{Reza Samavi}\Email{samavi@torontomu.ca}\\
\addr{Department of Electrical, Computer, and Biomedical Engineering, Toronto Metropolitan University, Toronto, ON, Canada\\Vector Institute, Toronto, ON, Canada}}

\begin{document}

\maketitle

\begin{abstract}
%Precise estimation of predictive uncertainty in deep neural networks is a critical requirement for reliable decision-making in machine learning and statistical modeling, particularly in the context of medical AI. Conformal Prediction (CP) has emerged as a promising framework for representing the model uncertainty by providing prediction sets associated with unseen data. However, producing well-calibrated uncertainty sets as confidence levels for individual predictions remains an active research area, yet to be fully addressed. 
%In this paper, we propose Evidential Conformal Prediction (ECP) that generates well-calibrated and precise uncertainty sets for unseen data using evidential learning and two subjective attributes associated with the target labels called Uncertainty Surprisal and Expected Utility. We propose an evidential approach to represent the model uncertainty by generating smaller and confident prediction sets that are guaranteed to have marginal coverage of true labels with a user-specified high probability. We explore state-of-the-art CP methodologies such as APS and RAPS, and compare them with our proposed method to show that ECP outperforms existing CP methods in terms of the average size and quality of produced prediction sets while maintaining marginal coverage with an arbitrary high probability. The source codes of the experiments are available at \url{}.   

In this paper, we propose \emph{Evidential Conformal Prediction (ECP)} method for deep
% it used to be image classifiers here before CR version
classifiers to generate the conformal prediction sets. Our method is designed based on a non-conformity score function that has its roots in Evidential Deep Learning (EDL) as a method of quantifying model (epistemic) uncertainty in DNN classifiers. We use evidence that are derived from the logit values of target labels to compute the components of our non-conformity score function: the heuristic notion of uncertainty in CP, uncertainty surprisal, and expected utility. Our extensive experimental evaluation demonstrates that ECP outperforms three state-of-the-art methods for generating CP sets, in terms of their set sizes and adaptivity while maintaining the coverage of true labels.
%enhances the model reliability by providing a formal guarantee in the model uncertainty boundaries. Furthermore, the method 
%facilitates the capability of CP to be compared with the existing uncertainty quantification approaches.
%Real-world applications can highly benefit from decision-making under uncertainty and risk assessment. This research advances model uncertainty quantification in conformal prediction, providing valuable approach for professionals, particularly, medical practitioners. Future research opportunities are identified for further development in this field.
\end{abstract}

\begin{keywords}
Conformal Prediction; Uncertainty Quantification; Evidential Deep Learning; Model Confidence; Deep Neural Networks;
\end{keywords}

\section{Introduction}
\label{intro}
Uncertainty Quantification (UQ) in Deep Neural Networks (DNNs) plays a crucial role in safety-critical applications, such as medical diagnosis~\citep{medical_cp} and robotics~\citep{robotics_cp}. Conformal Prediction (CP) or Conformal Inference~\citep{cp,inductive_cp} is a distribution-free, post-processing framework for UQ in image classification with promising properties such as no assumption on input data distribution, types of the pretrained model architecture, and domain of application. 
Rather than providing a single point prediction, CP constructs a finite \emph{prediction set} or \emph{uncertainty set} that encompasses a plausible subset of class labels for a given unseen input data point. 
%Suppose a radiologist tasked with diagnosing cancerous tumors from medical images. Conventional machine learning models might provide predictions with a certain level of confidence, e.g. softmax scores, but they lack guarantees about the accuracy of those predictions. This uncertainty could lead to critical misdiagnoses or unnecessary procedures. CP not only generates point predictions for each image, but also prediction sets that come with a guarantee in marginal coverage. This means that, for instance, if the radiologist aims for 90\% confidence in the predictions, they can be confident that at least 90\% of the true labels fall within the prediction sets in addition to considering other likely states included in the prediction sets. 
%When classifying a data point, in addition to a point estimation of the most likely predictive probability, 
In CP, three criteria are important to ensure the quality of the produced prediction sets: (1)~the inclusion of the true labels, with a high probability, 
%$1-\delta$, 
in the prediction set known as \emph{coverage}, (2)~the size of the set known as \emph{efficiency} so that larger the set size, higher the model uncertainty, and (3)~\emph{adaptivity} in both former properties: adaptivity in set size, i.e. producing relatively larger sets for difficult examples compared to easy examples (e.g., a well-projected image of a bird is considered easier to classify than a cropped one),   
%(having higher rank of true label's predictive probability) rather than the easy ones, 
and coverage, i.e. having higher probability for inclusion of true label in the prediction set. 
%smaller violation from $1-\delta$ in true label coverage. 
%Back to our motivating example, by providing statistically valid and efficient prediction sets, we can not only enhance diagnostic accuracy but also reduce unnecessary procedures, ultimately saving lives and resources.
A major challenge in CP research, is to construct high quality prediction sets where all these three properties are well maintained. 

%all criteria in practice, i.e. having smaller (more precise) and adaptive sets while maintaining the marginal coverage. 
To address this challenge, a \emph{non-conformity score} function is considered in CP to find an optimal probability threshold for inclusion of a label in the prediction sets. The score represents a measure of discrepancy between model predictive outcomes and true labels. The score is used to compare an unseen data point in a validation dataset with those in a relatively smaller calibration set (e.g., 500 examples) called the holdout dataset. The selection of this score function is arbitrary as long as it meets coverage property, yet the function plays a crucial role in meeting the other two properties when generating a prediction set: efficiency and adaptivity. 
For image classification tasks, two state-of-the-art (SoTA) CP methods are proposed with different non-conformity score functions, each addressing one or more CP properties. 
The \emph{Adaptive Prediction Set (APS)} approach~\citep{cp_aps} prioritizes adaptivity over efficiency by producing larger prediction sets (inefficient) using cumulative summation of sorted softmax probabilities in its non-conformity function. To further improve efficiency, \emph{Regularized Adaptive Prediction Sets (RAPS)} approach~\citep{cp_raps} regularizes APS's non-conformity scores to reduce the set size by penalizing the softmax probabilities for unlikely labels with two constant hyperparameters. 
% skipping the first $k_{reg}$ highest probable labels and penalizing the other softmax scores (for unlikely labels) by a fixed $\lambda$ as two hyperparameters.

In this paper, we propose \emph{Evidential Conformal Prediction (ECP)} as a CP method for image classifiers, with a non-conformity score function that better maintains CP properties compared to SoTA methods. Our approach has its roots in Evidential Deep Learning (EDL)~\citep{evid1} as a method of quantifying model (epistemic) uncertainty in DNN classifiers. We use this uncertainty as our heuristic notion of uncertainty and produce rigorous uncertainty using our proposed CP method. In our non-conformity function, the scores are computed based on the evidence derived from logit values of target labels. We map the evidence to belief masses, which are the probabilities indicating the amount of confidence on each label being the true label. The belief masses are used to compute our heuristic uncertainty. We also use evidence to compute parameters of a Dirichlet distribution that produce the predictive probabilities associated with the target labels. These predictive probabilities along with our heuristic uncertainty are used to compute two evidential properties: \emph{uncertainty surprisal} as the amount of information we still require to confidently predict the model outcomes, and \emph{expected utility} as a distribution over target labels expressing the performance or effectiveness of a learning model. Then, we define and compute our non-conformity score function using these evidential properties.
Our non-conformity scores are measurable without additional overhead. The coverage is theoretically guaranteed (\theoremref{cp_coverage_theorem}), and as per our extensive experimental evaluations, the average empirical coverage is stochastically the same as the user-specified coverage level while producing adaptive and smaller (more efficient) prediction sets compared to SoTA methods.
%% before CR version: 
% Our non-conformity scores are simple to measure and, as per our extensive experimental evaluation, guarantees coverage while producing adaptive and smaller (more efficient) prediction sets compared to all three SOTA methods.
%%%%%%

We are making three contributions: (1) we introduce ECP with a novel non-conformity score function leading to efficient prediction sets while maintaining their coverage and adaptivity, (2) we propose a reliability metric to measure the confidence and uncertainty associated with the expected coverage, i.e. marginal coverage over unseen input data, and (3) we propose a quality metric for the predication set based on violation from guaranteed coverage (as defined in~\citep{cp_raps}) and the average set size.

\section{Related Work}
\label{related_work}
In CP, there are SoTA methods to quantify model uncertainty either by generating a scaled value from the produced prediction sets to facilitate performing comparative evaluations with non-CP methods~\citep{uq_cp} or by improving the efficiency and adaptivity of prediction sets. 
A classical approach to generate prediction sets for unseen data is to include labels from the most likely to the least likely softmax probabilities until their cumulative summation exceeds the threshold $1-\delta$. In this approach, the true label coverage cannot be guaranteed since the output probabilities are overconfident and uncalibrated~\citep{calib2}. Furthermore, the lower probabilities in image classifiers are significantly miscalibrated which gives rise to larger prediction sets that may misrepresent the model uncertainty. There are also a few methods to generate prediction sets, but not based on conformal prediction~\citep{non_cp1,non_cp2}. However, these methods do not have finite marginal coverage guarantees as described in Theorem~\ref{cp_coverage_theorem}.
% can be used as input to a conformal procedure to potentially improve performance.

The coverage guarantee can be achieved using a new threshold and holdout data. In this regard, Romano et al.~\citep{cp_aps} proposed a method to make CP more stable in the presence of noisy small probability estimates in image classification. The authors developed a conformal method called APS to provide marginal coverage of true label in the prediction set which is also fully adaptive to complex data distributions using a novel conformity score, particularly for classification tasks. For example, with $\delta=0.1$, if selecting prediction sets that contain $0.85$ estimated probability can achieve $90\%$ coverage on the holdout data, APS will utilize the threshold $0.85$ to include labels in the prediction sets. However, APS still produces large prediction sets which cannot precisely represent the model uncertainty.
To mitigate the large set size, the authors in~\citep{cp_raps} introduced a regularization technique called RAPS to relax the impact of the noisy probability estimates which yield to significantly smaller and more stable prediction sets. RAPS modifies APS algorithm by penalizing the small softmax scores associated with the unlikely labels after temperature scaling~\citep{platt} using two hyperparameters $k_{reg}$ and $\lambda$. RAPS regularizes the APS method, therefore, RAPS acts exactly as same as APS when setting the regularization parameter $\lambda$ to 0. Both APS and RAPS methods are always certified to satisfy the marginal coverage in~\equationref{cp_coverage_eq} regardless of model, architecture, and dataset. Thus, RAPS could outperform APS by significantly smaller prediction sets. Thus, RAPS can produce adaptive but more efficient (smaller) prediction sets as an estimation of the model uncertainty given unseen image data samples.
% RAPS also performs better rather than choosing a fixed-size set. 
Both methods also require negligible computational complexity in both finding the proper threshold using the holdout data. However, both methods requires an optimization process to find optimal temperature and generate softmax scores as predictive probabilities. Moreover, RAPS requires another optimization process to find the optimal $\lambda$ that generates the most adaptive sets while sacrificing the set size. These further optimization process may give rise to higher computational burden in RAPS.
Another method to produce small sets is called Least Ambiguous Set-valued (LAS) introduced in~\citep{cp_lac}. LAS produces small average set size in the case where the input probabilities are correct. However, compared to SoTA methods, LAS has higher violations from the desired exact coverage probability $1-\delta$ conditional on ranges of set sizes.

There are also other methods to quantify model (epistemic) uncertainty in DNNs such as Bayesian~\citep{dropout,ensemble} and evidential~\citep{evid1, edl_reg} methods.~\citep{evid1} proposed the notion of Evidential Deep Learning (EDL) to quantify uncertainty in classification tasks and showed that their approach is robust for out-of-distribution testing data. EDL uses Dempster-Shafer Theory (DST) to produce evidence for target labels based on the logit values and maps the evidence to Dirichlet distribution parameters. EDL could effectively capture the epistemic uncertainty associated with classification tasks using Subjective Logic (SL)~\citep{sl} as a framework that formalizes DST. 
%They introduced a quantification method named DeepTrust, as a framework based on Subjective Logic in which the model opinion and trustworthiness are quantified. We also studied the limitations of this method (see Section~\ref{limit}). We used this propagation method to incorporate interpretable input data opinions and propose a comprehensive optimization process over opinions. 
In this paper, we exploit EDL originated in SL to devise our non-conformity score function in CP setting as described in the following sections. 
% SL as a framework is used to formalize 
  % In~\cite{hope}, the authors use SL to produce subjective opinions of data points to be propagated in a DNN model in order to obtain the total model opinion. 

% Consider a procedure that outputs a predictive set for each observation, and further suppose that this procedure has a tuning parameter τ that controls the size of the sets (In RAPS, τ is the cumulative sum of the sorted, penalized classifier scores). We take a small independent conformal calibration set of data, and then choose the tuning parameter τ such that the predictive sets are large enough to achieve 1 − α coverage on this set. This calibration step yields a choice of τ , and the resulting set is formally guaranteed to have coverage 1 − α on a future test point from the same distribution;

%\input{sections/background}
%\input{sections/method}
 \section{Evidential Conformal Prediction}
\label{cp_uncer}
In the following sections, we formally describe Evidential Deep Learning in~\sectionref{edl_back}, the procedure of computing evidential properties in~\sectionref{ecc_sec} as components of our proposed non-conformity score function leading to constructing efficient and adaptive prediction sets described in~\sectionref{evid_pred_set,insight_sec} which is introduced as \emph{Evidential Conformal Prediction (ECP)}. Finally, in~\sectionref{cov_conf}, we discuss the confidence and uncertainty associated with the true label coverage after producing prediction sets.

\subsection{Evidential Deep Learning}
\label{edl_back}
% \rs{(please avoid passive voice almost everywhere --is generalized--) generalizes}
% \rs {subjective (why subjective? you haven't defined what you mean by subjective!)} 
Dempster–Shafer Theory of Evidence (DST) generalizes Bayesian theory by individually assigning a probability called belief mass to each of the possible states of the model (e.g. model outcomes), thus, allowing DST to explicitly represent incomplete and partially reliable knowledge about the states.
Belief masses are probabilities that are assigned to subsets of a frame of discernment (including the entire frame itself) that denotes a set of exclusive possible states~\citep{dst,evid_math}.
Belief masses represent the amount of confidence that the truth can be any of the possible states. For example, in a DNN classifier, any combination of assigned target labels are possible states of the model outcomes as each class label can be assigned by a probability as a \emph{belief} mass to indicate the confidence on which class label is the true label.
%For example, in a multi-class classification task, each class label can be assigned by a probability as a belief mass to indicate the confidence that the class label is the true label when classifying an image. 
By assigning all belief masses to the entire possible model outcomes, the notion of “I do not know” can be expressed as the 
uncertainty that indicates lack of confidence over model outcomes~\citep{sl}.
%\rs {(this needs further clarification- base rate? prior - you should speak a language understandable for the community - e.g., make a connection between base rate and prior - you should not directly use Josang's language as he has pages and pages to develop the idea - here even that I dont know concept look immature)} 
The Dempster-Shafer theory is restricted by ignoring the base rates or prior probabilities associated with the possible outcomes, the theory of subjective logic formalizes and extends the DST’s notion of belief assignments to each of the model outcomes by the inclusion of the base rates associated with target labels and mapping the evidence obtained from input data to belief masses using a Dirichlet distribution. Therefore, in subjective logic, belief masses are function of evidence. For example, in a DNN classifier, evidence can be computed based on the logit values to form belief masses. 
Note that the notion of evidence we are using in this proposal, is different from what considered as marginal likelihood in Bayesian models. 
% \rs {evidence should be defined in terms of states and DST and BNN provide an example to connect it to classiofication task evidence }
% SL forms an informative structure called \emph{subjective opinion} about the model outcomes for an input data point~\cite{sl}. Opinions contain the quantified belief masses and uncertainty obtained from the predictive evidence over the model outcomes. 
% Unlike the Bayesian methods, SL can also define an ownership for the assigned belief distribution and its associated uncertainty.
% EDL exploits this mapping and evidence for the model outcomes considering their associated belief masses as an opinion to explicitly report the model uncertainty. 

Evidential Deep Learning built on Dempster-Shafer theory and subjective logic, obtains the evidence from the logit values in a DNN, and maps the produced evidence to Dirichlet parameters in order to form a predictive Dirichlet distribution. Informally, \emph{evidence} in evidential deep learning is the amount of support collected from input data in favor of classifying a data point into a specific label~\citep{evid1}.
%Different from the terminology used in Bayesian modeling for marginal likelihood, the term \emph{evidence} in EDL represents the amount of support collected from input data in favor of classifying a data point into a specific label~\citep{evid1}.
\begin{figure}[t]
\centering
\includegraphics[width=0.9\linewidth]{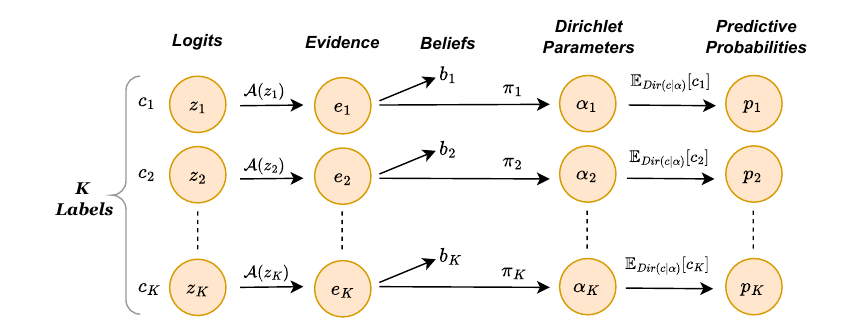}%
\caption{An overview of Evidential Deep Learning (EDL): For each data point, the evidence values are calculated based on logits to produce belief masses and Dirichlet parameters, and finally, achieving predictive probabilities.} % : (a) MNIST/NotMNIST, (b) F-MNIST/NotMNIST, (c) CIFAR10/CIFAR100, and (d) CIFAR10/SVHN
\label{overview}
\end{figure}
\begin{definition}[Evidence in DNN]
\label{evidence_def}
In a classifier with $K$ target labels, suppose $z_k$ is the logit value associated with the label $k \in \{1,2,...,K\}$ and $\mathcal{A}: \mathbb{R} \rightarrow \mathbb{R}^+$ is a non-linear non-negative activation function. Then, the non-negative real evidence $e_k$ associated with the label $k$ is defined as $e_k=\mathcal{A}(z_k)$.
\label{evidence_def}
\end{definition}
% \rs {extract a succinct definition from the paragraph below }
\begin{definition}[Base Rate (Prior Probability) in DNN]
\label{prior_def}
In a classifier with $K$ target labels, the prior probability $\pi_k$ associated with the label $k \in \{1,2,...,K\}$ is called base rate and defined as the probability of each label being the true label in classifying a data point prior to any observation or training process.
\label{base_rate_def}
\end{definition}
We illustrate the process of generating evidence, belief masses, and Dirichlet parameters from logit values in~\figureref{overview}. 
% \rs {(here you have to go and clearly explain everything through this figure, we start by labels ,,, blu blu all the way to uncertainty and define anything that you need)}
Formally, we replace the softmax layer with an activation layer (non-linear non-negative activation function $\mathcal{A}$, e.g. ReLU function), corresponding to a latent categorical distribution $\pmb{c}=\{c_1,c_2,...,c_K\}$ denoting the probabilities associated with target labels $\mathcal{Y}=\{1,2,...,K\}$. When classifying a data point, the activation function $\mathcal{A}$ is applied to logit values $z_k$ associated with target labels $k \in \mathcal{Y}$ to produce real non-negative evidence values denoted by $e_{k} \in \pmb{E}$ for each label $k$. Note that $\pmb{E}=\{e_{k}\ |\ k \in \mathcal{Y}\}$ denotes the set of evidence associated with target labels in $\mathcal{Y}$. 
% In the model $\mathcal{M}_\Theta$, to classify each data point $i$ to the label $k$ among $K$ target labels such that $k \in \mathbb{N}^{[1,K]}$, the evidence vector $E_i$ with $|E_i|=K$ is achieved from the model $\mathcal{M}_\Theta$ such that $\mathcal{A}(\mathcal{M}_{\Theta}(i))=E_i \succeq 0$. The evidence vector $E_i$ contains evidence $e_{ik} \in E_i$ associated with the target label $k$. 
Following~\definitionref{evidence_def} (evidence) and~\definitionref{base_rate_def} (base rate), now we can form a Dirichlet distribution with the parameter set $\pmb{\alpha}=\{\alpha_{k}\ |\ k \in \mathcal{Y}\}$ by computing its corresponding Dirichlet parameters $\alpha_{k} \in \pmb{\alpha}$ based on the corresponding evidence $e_k$ and the base rates $\pi_k$ as, 
%Then, as shown in~\figureref{overview}, a Dirichlet distribution $\pmb{\alpha}=\{\alpha_{k}\ |\ k \in \mathcal{Y}\}$ is formed by computing its corresponding Dirichlet parameters $\alpha_{k} \in \pmb{\alpha}$ based on the corresponding evidence $e_k$ and the base rates, i.e., prior probabilities $\pi_k$ as,
\begin{equation}
\alpha_{k} = e_{k} + K.\pi_k \quad \Longrightarrow \quad \alpha_{k} = e_{k} + 1\ ,
\label{dirich_parameter}
\end{equation} 
where the base rates $\pi_k$ were equally and uniformly selected as $\pi_k=\frac{1}{K}$ to represent equal prior probabilities associated with labels due to the ease of deployment and equal importance/priority among labels.
The Dirichlet distribution is a Probability Density Function (PDF) for possible labels of the latent categorical distribution $\pmb{c}$ that is denoted by $Dir(\pmb{c}|\pmb{\alpha})$. This PDF is used as the conjugate prior distribution for the categorical distribution $\pmb{c}$ since the posterior distribution after incorporating the knowledge gained from the observed data, is also a Dirichlet distribution. The Dirichlet distribution PDF $Dir(\pmb{c}|\pmb{\alpha})$ is characterized by the parameter set $\pmb{\alpha}$ as,
\begin{equation*}
Dir(\pmb{c}|\pmb{\alpha})=
\begin{cases}
    \frac{1}{B(\pmb{\alpha})}\prod_{i=1}^K c_i^{\alpha_i - 1} & \quad \text{for}\ 
 \pmb{c}\in \mathcal{V}_K \\
    0 & \quad \text{otherwise}
\end{cases}\ ,
\label{dir_pdf}
\end{equation*}
where $B(\pmb{\alpha})$ is the $K$-dimensional multivariate beta function~\citep{beta_func}, and $\mathcal{V}_K$ is the $(K-1)$-dimensional unit simplex such that
\begin{equation*}
\mathcal{V}_K=\{\pmb{c}\ |\ \sum_{i=1}^K c_i=1 \quad \text{and} \quad 0 \leq c_1,c_2,...,c_K \leq 1\}\ .
\label{simplex}    
\end{equation*} 
Therefore, the Dirichlet parameter set $\pmb{\alpha}$ is considered over all possible outcomes in classifying any given data sample. 
% Starting from prior distribution towards observing a data point, knowledge can then be updated based on the data point leading to a new distribution of the same form as the old one.
When a data sample is observed by the model, the sample is linked to one of the $K$ labels and the corresponding Dirichlet parameter $\alpha_k$ is increased, e.g., due to identifying a specific pattern within an image. Then, the Dirichlet distribution is updated based on the new observation. 
% \rs{(Uncertainty should also be defined as a definition inside the text as you havce also a formal definition of it as shown in this figure)}
The expected probability of the Dirichlet distribution, projected by each target label, is considered as the predictive probability $p_{k} \in \pmb{P}$ associated with the label $k$, and computed as,
\begin{equation}
\mathbb{E}_{Dir(\pmb{c}|\pmb{\alpha})}[c_k \in \pmb{c}] = p_{k}=\frac{\alpha_{k}}{\alpha_{0}}\ , \quad \text{where} \quad \alpha_{0} = \sum_{k=1}^{K} \alpha_{k}=K+\sum_{k=1}^K e_k\ .  
\label{prob}
\end{equation} 
% \begin{equation}
%  = \sum_{k=1}^{K} (e_{ik}+1) = K + \sum_{k=1}^{K} e_{ik}\ .
% \label{sum_parameter}
% \end{equation}
The summation of Dirichlet parameters denoted by $\alpha_0$ is called Dirichlet strength, indicating the total collected evidence in classifying a data point.
Using the Dirichlet distribution $\pmb{\alpha}$, the belief masses $b_{k}$ indicating the confidence associated with the label $k$, and the \emph{uncertainty} of the model (epistemic) are computed as,
\begin{equation}
b_{k} = \frac{e_{k}}{\alpha_{0}} \qquad \text{and} \qquad u = \frac{K}{\alpha_{0}} \qquad \text{s.t.} \qquad u + \sum_{k=1}^{K} b_{k} = 1\ ,
\label{belief}
\end{equation}
where $0\leq b_k<1$ denotes belief masses that are directly correlated with the evidence $e_k$ associated with the label $k$ and they are sub-additive to 1 (i.e. the summation of belief masses are less than 1). The model uncertainty $0<u\leq 1$ is considered as our heuristic notion of uncertainty and defined as follows:
\begin{definition}[Heuristic Model Uncertainty]
A complete lack of confidence over all target labels in a DNN classifier is called the model (epistemic) uncertainty over all possible model outcomes that is heuristically used to devise a non-conformity score function in CP. 
\label{uncer_def}
\end{definition}
The model uncertainty $u$ is inversely proportional to the total collected evidence for supporting the target labels. 
In CP, a non-conformity score function is devised based on a heuristic notion of model uncertainty. In our non-conformity score function, the model uncertainty $u$ obtained from evidential setting carries heuristic notion of uncertainty.

\subsection{Evidential Classification Cost}
\label{ecc_sec}
In CP, when we construct prediction sets, three distinct properties are required to be satisfied. The first property is the \textit{marginal coverage} or validity that guarantees to include the true label in the prediction sets based on the following theorem~\citep{cp}:
\begin{theorem}[Conformal Coverage Guarantee]
\ Consider $\mathcal{X}_{cal}$ and $(x_{val},y_{val}) \in \mathcal{X}_{val}$ are i.i.d. and correspond to unseen test data as holdout data and a validation data point, respectively. Let $\delta$ be the user-chosen coverage error level, $\widehat{q}$ is the $1-\delta$ quantile of holdout non-conformity scores, and $\mathcal{C}: \mathbb{R}^d \times \mathbb{R} \rightarrow 2^{\mathcal{Y}}$ be the set-valued predictor function for all possible labels in $\mathcal{Y}$. Then, the probability of the true label being covered in the prediction set is guaranteed as follows: 
\begin{equation}
   \mathcal{P}(y_{val} \in \mathcal{C}(x_{val},\widehat{q})) \geq 1-\delta\ .
\label{cp_coverage_eq}
\end{equation}
\label{cp_coverage_theorem}
\end{theorem} 
%%%%%%%%%%%%%%%% FULL THEOREM %%%%%%%%%%%%%%%%%%%%%%%%%%
% \begin{theorem}[Conformal Coverage Guarantee]
% \label{}
% \ Consider $\{(x_i,y_i) \in \mathcal{X}_{cal}\}_{i=1}^n$ and $(x_{val},y_{val}) \in \mathcal{X}_{val}$ are i.i.d. and unseen test data as $n$ holdout data and a validation data point, respectively. Let $\delta$ be the user-chosen coverage error level, $\widehat{q}$ is the $1-\delta$ quantile of calibration conformal scores, and $\mathcal{C}: \mathbb{R}^d \times \mathbb{R} \rightarrow 2^{\mathcal{Y}}$ be the set-valued predictor function. If $\mathcal{C}(x_{val},\widehat{q})$ gradually grows to include all possible labels in $\mathcal{Y}$ when having large enough $\widehat{q}$, then, the probability of the true label being covered in the prediction set is guaranteed in the following bounds: 
% \begin{equation}
%     1-\delta \leq \mathcal{P}(y_{val} \in \mathcal{C}(x_{val},\widehat{q})) \leq 1-\delta+\frac{1}{n+1}\ .
% \label{cp_coverage_eq}
% \end{equation}
% \label{cp_coverage_theorem}
% \end{theorem} 
% \noindent The proof and the related conditions of this theorem are available in~\citep{inductive_cp}. % \cite{cp,distribution_cp}
The second property is the \textit{set size} or efficiency to reflect the desirability of a smaller size for the  prediction set.
The third property is the \textit{adaptivity} that necessitates the set size for unseen data is modified to represent instance-wise model uncertainty, i.e., the set size is smaller when the model encounters easier test data rather than the inherently harder ones. Note that the difficulty of a test data point is based on the rank of its true label in the sorted set of outcome probabilities. 
These properties affect each other; for example, the set size property tries to make the sets smaller, while the adaptivity property tries to make the sets larger for harder data points when the model is uncertain. Similarly, choosing sets with a fixed size may satisfy the coverage property, but without adaptivity.
%\subsection{Evidential Properties of Target Labels}
%\label{evid_prop}

In multi-class classification tasks, a DNN classifies a data point to a specific target label; the one with the highest predictive probability. In this process, not all unselected target labels for the given data point are equivalent as each one of them  may carry a different predictive probability, albeit the probability is lower than the probability of the selected target label. We use this notion along with other evidential properties to define \emph{Evidential Classification Cost (ECC)} for each label $k \in \mathcal{Y}$ as follows:
%This classification process is accompanied by a cost which can be subjectively assigned to each target label . We define the concept of \emph{Evidential Classification Cost (ECC)} as follows:
\begin{definition}[Evidential Classification Cost]
The amount of cost or error that a pretrained classifier indicates based on evidential properties when classifying a data point into a specific target label. 
\end{definition}
In other words, ECC represents an estimation of how much each target label $k \in \mathcal{Y}$ is not a correct choice to classify a particular data point.  
The evidential properties are used as additional information to support not only for the true label, but also for all other labels in each prediction. 
Thus, a model produce the lowest value (minimum) of ECC for the true label when correctly classifying a data point.
% Each label’s attributes are exploited and measured to show the trustworthiness of the classification into each label~\citep{sl}. 
% This subjectivity comes from two specific properties of opinions in subjective logic~\cite{sl}: 1) the evidence and beliefs are owned and affected by individuals, i.e., each target label in the learning model, and 2) the beliefs are influenced by uncertainty associated with the model outcomes. 
We use two evidential properties \emph{Uncertainty Surprisal} and \emph{Expected Utility} to subjectively quantify ECC associated with each target label leading to computing our non-conformity score function. 
We call this cost subjective as it reflects the notion that these two properties are carried and affected by each one of the target labels individually. 
ECC is different and computationally independent from the objective cost computed by the model's loss function (objective function). 
% The subjectivity, as described in SL, 
%We should emphasize that the list of evidential properties introduced here is neither complete nor exhaustive.
% If such evidential properties associated with the target labels exist, one can force the model to produce a significantly lower amount of evidence and, therefore, higher model uncertainty when misclassifying an ID data point or encountering OOD data. 

% \rs{(I thinkl you're not consistent with your use of Definitions, why not use Def for Uncertainty surprisal or expected utility or evid pred set?)}
% \vspace{2mm}
% \noindent \textbf{Uncertainty Surprisal.}

According to~\equationref{prob,belief}, the model uncertainty $u$ is achieved by considering the beliefs associated with all the target labels when classifying a data point. The portion of the uncertainty $u$ associated with each label $k$ is called \emph{Focal Uncertainty} $U_{k}$ that is partitioned based on the corresponding prior probability $\pi_{k}$ as,
\begin{equation}
U_{k} = u\pi_{k}\ .
\label{fu}
\end{equation}   
As we obtain the focal uncertainty associated with each label, we can now measure the first evidential property called uncertainty surprisal that is defined in information theory as follows~\citep{sl}:
\begin{definition}[Uncertainty Surprisal]
The uncertainty surprisal distribution $I_U$ over target labels is the amount of information required to confidently predict the model outcomes based on its corresponding predictive probability distribution.
\end{definition}
In EDL, the predictive probability $p_{k}$ is considered as the projected probability for the target label $k \in \{1,2,...,K\}$ when the pretrained model $\mathcal{M}$ is tested with each data point. The surprisal of the target label $k$ is then computed by the negative logarithm of the predictive probability $p_{k}$ as,
\begin{equation}
I(k) = - \log p_{k}\ .
\label{surprisal}
\end{equation}
The surprisal function, $I(k)$, measures the degree of an outcome is surprising. High surprisal shows that we require more information to predict the outcome,  %resulting in a less likely outcome, 
whereas low surprisal shows that we require less information to predict the outcome.
%resulting in a more likely outcome. 
Therefore, the uncertainty is directly intertwined with the surprisal such that the more surprisal associated with a target label, the more uncertainty we encounter in selecting the label to classify an input data point. 
To avoid misclassification and maintain the performance, we expect the model to have low surprisal on a specific target label. As we aim to produce adaptive non-conformity scores for target labels of each data point, we exploit a particular surprisal function based on the focal uncertainty associated with each target label. This surprisal function is called \emph{Focal Uncertainty Surprisal} which is denoted by $I_U(k)$ for a data point and defined as,
\begin{equation}
I_U(k) = \frac{U_{k} I(k)}{p_{k}}\ ,
\label{fu_surprisal}
\end{equation}
where $p_{k}$ is the predictive probability, $I(k)$ is the surprisal function, and $U_{k}$ is the focal uncertainty associated with the label $k$. Note that as shown in~\equationref{fu}, the model uncertainty $u$ and  the base rate $\pi_{k}$ (i.e., prior probability) of the label $k$ are taken into account to compute focal uncertainty $U_{k}$. 
Now, we define the second evidential property called expected utility as follows~\citep{sl}:
% \vspace{2mm}
% \noindent \textbf{Expected Utility.}
\begin{definition}[Expected Utility]
Considering the predictive probability distribution $\pmb{P}$ over possible model outcomes, an expected utility distribution $\Phi$ over target labels is a distribution expressing the expected performance or effectiveness of a learning model associated with the model outcome distribution.
\end{definition}
In a learning model, we compute the expected utility distribution associated with target labels by the element-wise product of $\pmb{\varphi}$ over target labels and the corresponding predictive probability distribution $\pmb{P}$ obtained by Dirichlet parameters. Specifically, the expected utility $\Phi(k)$ of the model $\mathcal{M}_{\Theta}$ when classifying a data point to the label $k$ is defined as,
\begin{equation}
\Phi(k) = \phi_{k} p_{k}\ ,
\label{exp_utility}
\end{equation}   
where $\phi_{k} \in \pmb{\varphi}$ and $p_{k} \in \pmb{P}$ denote the utility and the predictive probability associated with the target label $k$ when a data point is classified.
Since softmax probabilities can represent the performance of a learning model when classifying a data point into a target label, we apply softmax function $\sigma(.)$ to the $K$ logits and consider it as the corresponding utility distribution $\pmb{\varphi}$ when classifying the data point $x \in \mathcal{X}$ such that $\pmb{\varphi} = \sigma(\mathcal{M}(x))$. Note that it is also possible to consider other aspects of the model performance or effectiveness as the utility distribution associated with the model outcomes.

\subsection{Evidential Prediction Set}
\label{evid_pred_set}
For a pretrained model, we use \emph{split conformal prediction} in which we split all the available test (unseen) data $\mathcal{X}$ into holdout (calibration) data $\mathcal{X}_{cal}$ and validation data $\mathcal{X}_{val}$ such that $\mathcal{X}=\mathcal{X}_{cal} \cup \mathcal{X}_{val}$ and $\mathcal{X}_{cal} \cap \mathcal{X}_{val}=\emptyset$. When classifying an unseen data point (e.g., an image), the difficulty level of the data point denoted by $r_{y}$ is defined as the rank of its true label $y \in \mathcal{Y}$ in the descending-sorted set of predictive probabilities from the highest to lowest values denoted by $\langle p_k \rangle$~\citep{cp_raps}. This difficulty level can be incorporated in ECC and non-conformity score since the prediction sets should be larger for more difficult images that has higher true label rank rather than the easier ones due to adaptivity criterion. Therefore, in classifying a data point, when the rank of label $k$ denoted by $r_k \in \{0,1,...,K-1\}$ in a descending-sorted set of predictive probabilities $\langle \pmb{P} \rangle$ is higher, the ECC associated with the label $k$ is also increased. We define the scaling function $\rho(k)$ to quantify the impact of $r_k$ on ECC as,
\begin{equation}
\rho(k) = \frac{K}{K-r_k}\ ,
\label{rank_ecc}
\end{equation}
where $K$ is the constant number of possible target labels and $r_k$ is the rank of label $k$ in descending-sorted set of predictive probabilities. To incorporate $r_k$ in ECC, we require $\rho(k)$ to scale $r_k$ and suppress its extent. Note that $\rho(k)$ is an increasing function with respect to increasing $r_k$. Furthermore, when $r_k=0$ (i.e., first rank), $\rho(k)$ has no impact on ECC, the label $k$ is the most probable label and have the highest predictive probability as $p_k=\underset{\nu \in \mathcal{Y}}{\max}\ p_{\nu}$.

\vspace {.1cm}
We denote ECC associated with the label $k$ as $C_k$ that is dependent on the two evidential properties (i.e., uncertainty surprisal and expected utility) and the rank associated with the target labels. Following the interpretations of focal uncertainty surprisal in~\equationref{fu_surprisal} and expected utility in~\equationref{exp_utility}, ECC is directly proportional to the focal uncertainty surprisal as $C_k \propto I_U(k)$, and inversely proportional to the expected utility as $C_k \propto \frac{1}{\Phi(k)}$. Lower predictive probability $p_k$ gives rise to higher focal uncertainty surprisal $I_U(k)$, lower expected utility $\Phi(k)$, higher entropy, higher model uncertainty $u$, and ultimately, higher $C_k$ for label $k$ in classifying a data point. According to~\equationref{rank_ecc}, ECC is also directly proportional to the label rank $r_k$ and its scaling function $\rho(k)$ as $C_k \propto \rho(k)$ since higher rank for label $k$ implies lower $p_k$ that gives rise to higher $C_k$. Using~\equationref{rank_ecc}, the difficulty of a data point, i.e. true label rank, can contribute to both the proposed $C_k$ and non-conformity score $S_{ECC}(k)$. Thus, we define evidential classification cost $C_{k}$ associated with classifying a data point into the target label $k$ as,
% \frac{-\frac{a_{k}.u_i.\log p_{ik}}{p_{ik}}}{\lambda_{ik}.p_{ik}}
\begin{align}
\begin{split}
        &C_{k} = \frac{\rho(k)I_U(k)}{\Phi(k)} = Ku \overbrace{\frac{-\pi_{k}\log p_{k}}{\phi_{k} p_{k}^2(K-r_k)}}^{\widehat{C}_{k}}\\ \Longrightarrow \qquad &C_{k} = Ku\widehat{C}_{k} \qquad \text{such that} \qquad \widehat{C}_{k}=-\frac{\pi_{k}\log p_{k}}{\phi_{k} p_{k}^2(K-r_k)}\ , 
\end{split}
\label{subjective_cost_prop}
\end{align}
where $K$ is the number of possible labels, $u$ denotes the model uncertainty, $p_{k}$ denotes the predictive probability, $\pi_{k}$ denotes the base rate, $r_k$ denotes the rank of the label $k$'s probability. $\phi_{k} = \sigma_{k}(\mathcal{M}(x))$ denotes the utility function associated with the data point $x~\in~\mathcal{X}$ and the target label $k$ in the model $\mathcal{M}$ which is considered as the softmax probability of the label $k$ (note that $\sigma(.)$ is the softmax function). We also call the term $\widehat{C}_{k}$ in~\equationref{subjective_cost_prop} as \emph{Label Evidential Cost}, which is completely dependent on the label $k$. 
% as,
% \begin{equation}
% \widehat{C}_{ik} = -\frac{a_{k}.\log p_{ik}}{\lambda_{ik}.p_{ik}^2} \quad \text{s.t.} \quad \lambda_{ik} = \sigma_{k}(\mathcal{M}_{\Theta}(i))\ ,
% \label{label_subjective_cost}
% \end{equation}
% where $\sigma_{k}(\mathcal{M}_{\Theta}(i))$ denotes the softmax probability corresponding to the label $k$ when the input data point $i$ is classified by the model $\mathcal{M}_{\Theta}$.
Then, for each data point $x \in \mathcal{X}$, we normalize (scale) $C_k$ to define the proposed evidential non-conformity score $S_{ECC}(k)$ associated with label $k$ that is computed as, 
\begin{equation}
\forall k \in \mathcal{Y}: \quad S_{ECC}(k) = \frac{C_k}{\underset{\nu \in \mathcal{Y}}{\max}\ C_{\nu}}\ ,
\label{ecc_cp_score}
\end{equation}
where $C_k$ is ECC associated with the label $k \in \mathcal{Y}$ for a single data point. 

To construct our evidential prediction set $\mathcal{C}(x_{val},\widehat{q})$ for an unseen data point $x_{val} \in \mathcal{X}_{val}$, we first apply the proposed evidential non-conformity score $S_{ECC}(k)$ defined in~\equationref{ecc_cp_score} to the data in holdout set $\mathcal{X}_{cal}$ that is uniformly and randomly partitioned from testing dataset such that $|\mathcal{X}_{cal}|=n$. Then, for each holdout data point, we order the evidential scores $S_{ECC}(k)$ for all $k \in \mathcal{Y}$ based on the descending-sorted set of their predictive probabilities $\langle p_k \rangle$, i.e. evidential scores associated with labels having the highest to the lowest predictive probabilities. We can now select the evidential non-conformity score $S_{ECC}(k=r_y)$ associated with the rank of the true label $y \in \mathcal{Y}$ in the descending-sorted set of predictive probabilities. Once we have $n$ holdout evidential scores corresponding to holdout data, we quantify the $(1-\delta)$-quantile of these scores as $\widehat{q}$. 
% Practically, we use the slightly rectified version of this quantile by quantifying $\frac{\lceil(n+1)(1-\delta)\rceil}{n}$-quantile of the evidential scores to mitigate the finite sample effect on marginal coverage by the limited holdout set. 
The quantile $\widehat{q}$ acts as a threshold that determines for each unseen data point, which one of the labels should be included in the prediction set $\mathcal{C}$. Finally, for each unseen data point $x_{val} \in \mathcal{X}_{val}$, we calculate $S_{ECC}(k)$ in~\equationref{ecc_cp_score} for all the labels $k \in \mathcal{Y}$ exactly as the same procedure we had for holdout data, and include each label $k$ that meets the following inequality in the prediction set $\mathcal{C}(x_{val},\widehat{q})$: 
\begin{equation}
\mathcal{C}(x_{val},\widehat{q}) = \{\ k \in \mathcal{Y}\ |\ S_{ECC}(k) \leq \widehat{q}\ \}\ ,
\label{ineq_cp}
\end{equation}
where $\widehat{q}$ is $(1-\delta)$-quantile of the evidential non-conformity scores $S_{ECC}(k=r_y)$ associated with the true label rank $r_y$ of holdout data $\mathcal{X}_{cal}$.

\subsection{Insights on Evidential Conformal Prediction}
\label{insight_sec}
According to evidential non-conformity score $S_{ECC}(k)$ associated with labels $k \in \mathcal{Y}$ introduced in~\equationref{ecc_cp_score}, we have several insights that shows ECP is a promising method in producing prediction sets in CP: (1) When the predictive probability $p_k$ is decreased due to misclassification or out-of-distribution input data, $C_k$ is increased significantly due to the different components containing $p_k$ in~\equationref{subjective_cost_prop} and consequently, $S_{ECC}(k)$ will be increased which represents higher uncertainty and worse fit between $x$ and the label $k$. (2) Based on~\equationref{belief}, as more evidence is collected when classifying a data point, i.e. $\sum_{k=1}^K e_k$, the model uncertainty $u$ associated with all the labels is decreased that yields to decreasing $C_k$ and $S_{ECC}(k)$ for the label $k$. Therefore, the model uncertainty $u$ in~\equationref{subjective_cost_prop} acts as a coefficient to further adapt the non-conformity score with the input data, particularly when the model encounters an input far from the training distribution.
(3) In~\equationref{subjective_cost_prop}, the base rate or prior probability $\pi_k$ contributes to computing $C_k$ (ECC) associated with the target label $k$. Since changing the base rate of each target label can lead to a change in the $C_k$ of the target label, prior probabilities are incorporated in the ECC which can be considered as a parameter to consider fairness or balance class frequency among possible classes in the proposed non-conformity score function. (4) The choice of utility function is not limited to the softmax function. A modeller can consider other measurements or functions to produce utility distribution $\pmb{\varphi}$ associated with labels, but with respect to the correlation and inverse proportionality between $\Phi(k)$ and $C_k$. (5) Using evidential setting, we can measure the reliability of marginal coverage by computing the uncertainty and confidence associated with coverage as described in the next section. 

%%%%%%%%%%%%%%%%%%%%%%%%%%%%%%%%%%%%%%%%%%%%%%%%%%
%%%%%%%%%%%%%%%%%%%%%%%%%%%%%%%%%%%%%%%%%%%%%%%%%%

% In conformal procedure, the final result is a prediction set that contains a subset of all possible predictive labels given an unseen data point in a pretrained model. This prediction set is normally considered as an indicator of uncertainty such that larger set size represents higher model uncertainty.  

\subsection{Reliability in Marginal Coverage}
\label{cov_conf}
According to stochasticity and randomness in sampling the holdout set with size $|\mathcal{X}_{cal}|=n$ in CP, different trials on infinite number of validation data indicate different marginal coverage in prediction sets. Based on~\equationref{cp_coverage_eq}, the marginal coverage is at least $1-\delta$ over this random sampling of holdout set. However, for a fixed holdout set, the marginal coverage on infinite validation data is not exactly $1-\delta$.  
To restrict the deviations from $1-\delta$ in the marginal coverage, we require to use holdout set with adequately large $n$ while having the coverage distribution~\citep{cp_book}. Formally, the distribution of coverage given a fixed holdouts set follows a Beta distribution~\citep{cp_conditional} based on the size of holdout set $n$ as,
\begin{align}
\begin{split}
\mathcal{P}(y_{val} \in \mathcal{C}(X=x_{val})\ |\ \mathcal{X}_{cal}) &\equiv \mathbb{E}\big[y_{val} \in \mathcal{C}(X=x_{val})\ |\ \mathcal{X}_{cal}\big]\sim Beta(n+1-l, l)\\ &\text{s.t.} \qquad l=\lfloor(n+1)\delta\rfloor\ ,
\end{split}
\label{cov_dist}
\end{align} 
where $n=|\mathcal{X}_{cal}|$ denotes the size of holdout set, and $\delta$ denotes the coverage error level.
This conditional expectation on fixed holdout data represents the coverage associated with a large set of validation data (infinite validation data in its perfect form).  

Following~\theoremref{cp_coverage_theorem}, consider $\delta$ as the error level of true label coverage, $|\mathcal{X}_{cal}|=n$ is the size of holdout set, and $\mathcal{C}:\mathbb{R}^d \rightarrow 2^{\mathcal{Y}}$ as the prediction set function for unseen validation data $x_{val}\in \mathcal{X}_{val}$ with the true label $y_{val}$. As the coverage distribution is conditional on a fixed holdout set, the uncertainty in coverage and the confidence associated with correctly covering the true label is highly dependent on $n$. 
We propose the following theorem to quantify the confidence and uncertainty associated with true label coverage as the reliability measures:

\begin{theorem}[Coverage Confidence Guarantee]
Suppose $\mathcal{Y}$ be the set of target labels, $\mathcal{X}_{cal}$ be the holdout set with size $n$, and $\mathcal{C}: \mathbb{R}^d \rightarrow 2^{\mathcal{Y}}$ be the set-valued predictor with the coverage error level $\delta$. If for infinite number of validation data $\mathcal{X}_{val}$, the expected true label coverage of $\mathcal{C}$ is conditional on a fixed holdouts set $\mathcal{X}_{cal}$, then $\mathcal{C}(x_{val}\in \mathcal{X}_{val})$ is $\gamma$-confident in coverage such that:
\begin{equation}
\gamma = \frac{n-\lfloor(n+1)\delta\rfloor}{n+1} \qquad \text{where} \qquad
\frac{n}{n+1}-\delta \leq \gamma < 1-\delta\ ,
\label{}
\end{equation}
and the uncertainty $U_\mathcal{C}$ in the expected coverage of $\mathcal{C}$ is computed as,
\begin{equation}
U_\mathcal{C} = \frac{2}{n+1}\ .
\label{}
\end{equation}
\label{conf_cov_theorem}
\end{theorem}
The proof of~\theoremref{conf_cov_theorem} is provided in Appendix~\ref{proof_append}.
This theorem indicates that with a large number of validation data, the expected coverage of true labels in prediction sets given a fixed set of holdout set is associated with an amount of uncertainty $U_\mathcal{C}$. Thus, we consider $\gamma$ degree of confidence in covering the true labels in prediction sets. A modeller can quantify the reliability in conformal coverage using these two measures. In the following section, we elaborate on our practical experiments in detail and demonstrate the evaluation results for the proposed method compared to the SoTA CP methods in producing prediction sets.

\section{Experimental Evaluations}
\label{exp}
% \rs {Look Hamed, you are only a week away from an extended deadline and you still do not have a solid writing of your results which apparantly needs at least 10 iterations to be completed. and if I didn't push you - you were still playing with your data! you do the easier part of coding and playing with data and postpone the hard  part of writing - you rush at the end and you have seen the outcome.  - THis is not the way it works - but I haven't been able to convey this notion to you in the past three years! this two sections should have been ready at least a month ago. Yet we are here with a weak background - not a complete methodology and an unwritten results and discussion a week before the submission! you rush to write confusing stuff in last minutes perhaps to avoid my critiques and.... You do a great job and then you ruin it with a bad articulation and presentation.}

In this section, we report our experiments on the validity (true label coverage), efficiency (set size), and adaptivity for the produced prediction sets of our approach \emph{ECP} compared to two state-of-the-art (SoTA) conformal approaches, \emph{APS}~\citep{cp_aps}, and \emph{RAPS}~\citep{cp_raps}, along with a baseline approach as \emph{Base} method. 
The \emph{Base} approach directly utilizes softmax probabilities to include all predicted labels from the most likely to the least likely in the prediction set until their cumulative probabilities exceeds a user-specified probability threshold.
%$1-\delta$, but 
Note that in \emph{Base} approach, true label coverage may not be guaranteed.

\vspace{1mm}
\noindent \textbf{Model Architectures and Datasets.} For fair comparative evaluations with SoTA methods, we are using nine pretrained image classifiers: ResNet~\citep{resnet}, VGG~\citep{vgg}, DenseNet~\citep{densenet}, ShuffleNet~\citep{shufflenet}, Inception~\citep{inception}, and ResNeXT~\citep{resnext} from Pytorch framework with standard hyperparameters. 
% in order to indicate scalability of the proposed method.
% \rs {(here you're describing your arch and dataset: this part doesn't belong here)}  
We also used two different test sets of ImageNet with 1000 class labels \emph{ImageNet-Val} (with 50K test images) and \emph{ImageNet-V2} (with 10K test images) after standard transformations and normalization. 
%These ImageNet test sets are adequately complicated and challenging (having a large set of target labels) for image classifiers for the purpose of assessments. 

\vspace{1mm}
\noindent \textbf{Experimental Setup.} We implemented our method in Python using PyTorch framework\footnote{The source codes of the experiments are available at: \url{https://github.com/tailabTMU/ECP}}, and the assessments have been performed on a machine with Intel(R) Core(TM) i7-10750H 2.60GHz (6 cores) processor, NVIDIA Tesla T4 GPU, and 32GB RAM allocated to our computation. 
% Unlike ECP that does not rely on softmax scores as labels' probabilities (see~\sectionref{edl_back}), 
We apply temperature scaling~\citep{calib1} for Base, APS, and RAPS methods using holdout data to find the optimal temperature for calibrating softmax scores before using them as predictive probabilities. In our experiments, we use 15K and 3K of data as holdout set to calibrate and quantify the $(1-\delta)$-quantile $\widehat{q}$ of non-conformity scores for ImageNet-Val and ImageNet-V2, respectively. We evaluate RAPS using its regularization hyperparameters $k_{reg}=5$ and different choices of constant penalty called $\lambda=\{0.0001,0.001,0.01,0.1,1\}$ that were introduced to encourage small set sizes by skipping the first $k_{reg}$ highest probable labels ($0\leq k_{reg}\leq K$) and then, penalizing the other softmax scores by a fixed $0<\lambda \leq 1$. Note that since RAPS is the regularized version of APS, we represent APS by setting $\lambda=0$ in RAPS. We conduct four sets of experimental evaluations as reported in the forthcoming subsections. Note that all results are averaged over different trials, and the negligible discrepancies between the empirically reported and theoretically guaranteed (\theoremref{cp_coverage_theorem}) coverage results are due to stochasticity and randomness in finite number of trials.

\subsection{Experiment 1: Marginal Coverage and Prediction Set Size}
\label{cov_size_exp}
\begin{figure}[t] 
\centering
\subfigure[Prediction Set Size]{
    \includegraphics[width=0.47\linewidth]{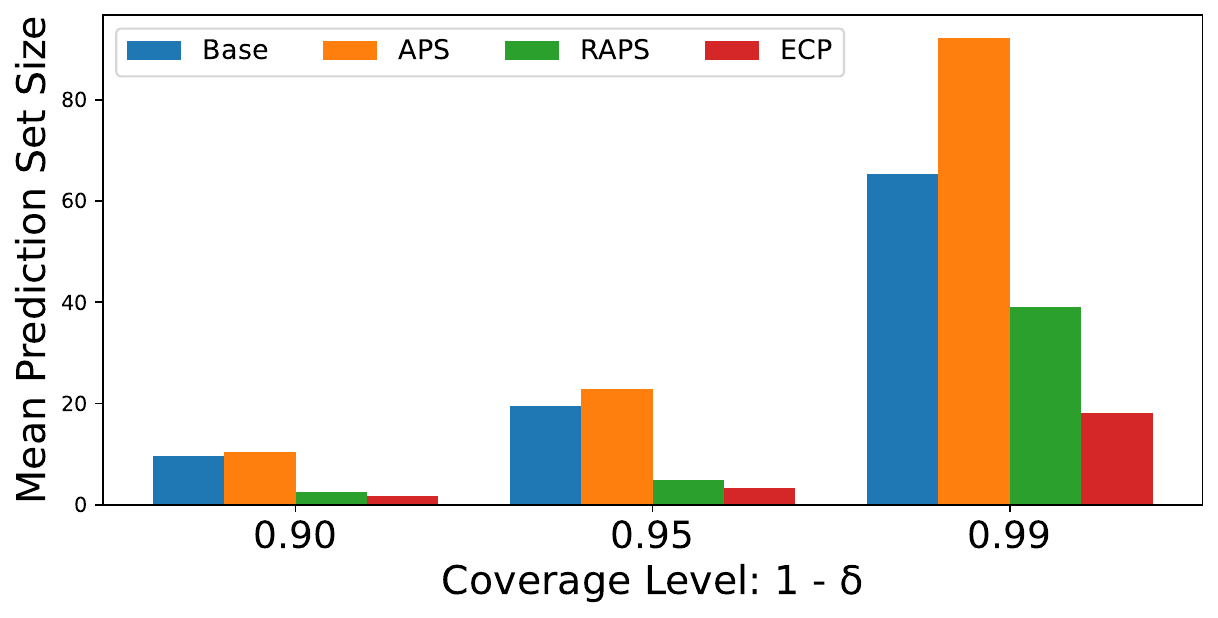}%
    \label{ecp_size}
    }
~
\subfigure[Marginal Coverage]{
    \includegraphics[width=0.47\linewidth]{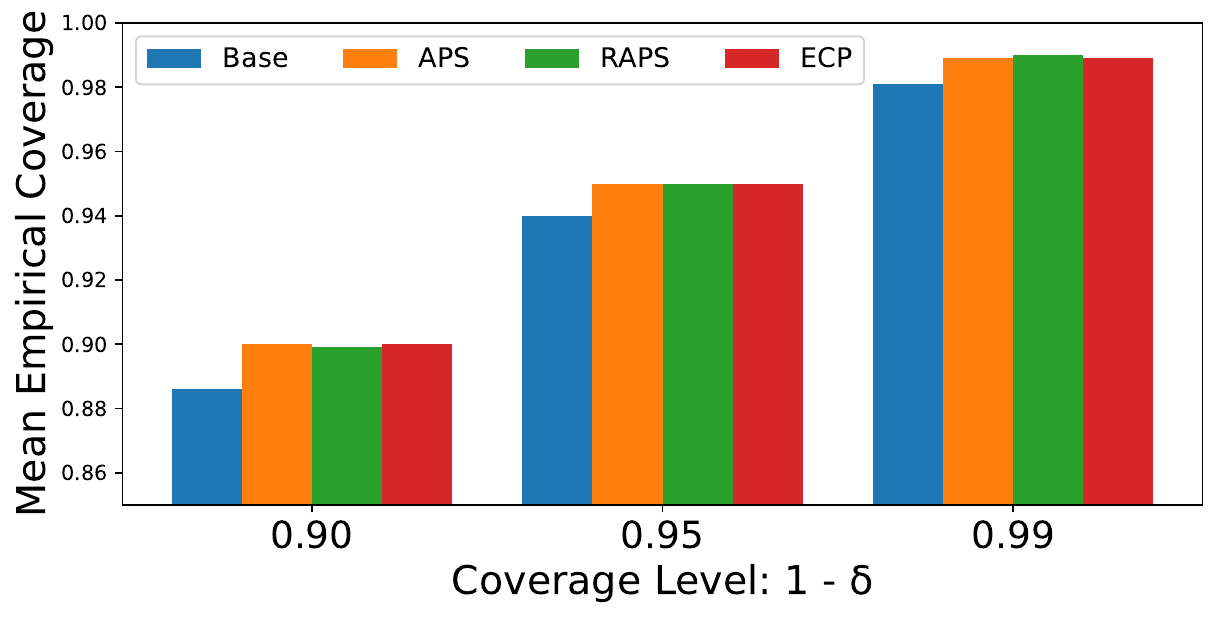}%
    \label{ecp_cov}
    }
\caption{Comparing ECP with SoTA methods using ResNet-152 with different choices of $\delta=\{0.1,0.05,0.01\}$ in terms of the average marginal coverage and prediction set size. For RAPS, we used $k_{reg}=5$ and $\lambda=0.1$.}
\label{ecp_comp}
\end{figure}

In this experiment, we calculate the average marginal coverage and set size of our method and each of the other three SoTA methods for two different choices of coverage error levels $\delta=0.1$ reported in~\tableref{perform_imagenetval} and $\delta=0.05$ reported in Appendix~\ref{size_append} over 10 different trials. We use both ImageNet-Val and ImageNet-V2 test data, and in each trial, we randomly split the test data into two subsets as 30\% of holdout set (calibration set) and 70\% of validation set. 
%as described in Experimental Setup. 
We report the mean over trials for average marginal coverage and set size for ImageNet-Val and ImageNet-V2 in~\tableref{perform_imagenetval,perform_imagenetv2}, respectively. We also plot the results of ResNet-152 model as our base classifier on ImageNet-Val to compare the performance of our method with the other CP methods when choosing different coverage error level in Figure~\ref{ecp_comp}. 
Unlike RAPS, ECP does not require any hyperparameters to be fine-tuned for a specific purpose which may cause computational burden. Therefore, we use $k_{reg}=5$ and $\lambda=0.1$ in these experiments as benchmark reported in~\citep{cp_raps} to apply RAPS method. 

\begin{table}[t]
\centering
\caption{Efficiency on \textbf{ImageNet-Val}: The median-of-means for marginal coverage and prediction set size using different architectures and methods with $\pmb{\delta=0.1}$ over 10 trials.}
\scalebox{0.85}{
\begin{tabular}{l c cccc ccc>{\bfseries}c}
    \toprule
\multirow{0}{*}{\textbf{Model}} 
        & \multicolumn{1}{c}{\textbf{Performance}} & \multicolumn{4}{c}{\textbf{Mean Marginal Coverage}} &
        \multicolumn{4}{c}{\textbf{Mean Prediction Set Size}} \\
    \cmidrule(lr){2-2} \cmidrule(lr){3-6} \cmidrule(lr){7-10}
        & Accuracy & Base & APS & RAPS & ECP & Base & APS & RAPS & ECP (Ours) \\
    \midrule
ResNeXT101 
        & 79.31\%  & 0.887 & 0.900 & 0.900 & 0.900 & 17.76 & 20.62 & 2.63 & 1.64 \\
    \addlinespace
ResNet152
        & 78.31\%   & 0.893 & 0.901 & 0.900 & 0.900 &  9.64 & 10.36 & 2.67 & 1.75 \\
    \addlinespace
ResNet101
        & 77.37\%  & 0.897 & 0.900 & 0.900 & 0.900 & 10.19 & 10.81 & 2.75 & 1.85 \\
    \addlinespace
ResNet50
        & 76.13\%   &  0.895   &  0.900  &  0.900  & 0.901 &  11.64 & 12.93 & 3.05 & 2.04    \\
    \addlinespace
ResNet18
        & 69.76\%   &  0.896   &  0.900  &  0.899  & 0.900 &  15.60 & 16.53 & 4.38 & 3.72    \\
    \addlinespace
DenseNet161
        & 77.13\%   &  0.892   &  0.901  &  0.899  & 0.901 &  11.71 & 12.82 & 2.83 & 1.82    \\
    \addlinespace
VGG16
        & 71.59\%   &  0.895   &  0.901 &  0.901 & 0.900 &  13.81 & 14.16 & 3.70 & 2.97    \\
    \addlinespace
ShuffleNet
        & 69.36\%   &  0.896   &  0.901  &  0.900  & 0.899 &  29.33 & 32.40 & 4.56 & 3.98    \\
    \addlinespace
Inception
        & 69.54\%   & 0.886   &  0.901  &  0.900  & 0.900 &  76.00 & 89.84 & 4.99 & 4.18    \\
    % \addlinespace

    \bottomrule
\end{tabular}  }
\label{perform_imagenetval}
\end{table}

\vspace{1mm}
\noindent \textbf{Observations.} We observe that our method (ECP) outperforms other methods in terms of prediction set size (last column) while choosing different coverage levels $1-\delta$ in~\figureref{ecp_size}. Furthermore, as shown in~\figureref{ecp_cov}, ECP guarantees the marginal coverage with different $\delta$ as same as SoTA methods except Base approach due to the noise in tail softmax probabilities.
Based on the results reported in~\tableref{perform_imagenetval,perform_imagenetv2}, for each model architecture, ECP (the boldfaced column) has always smaller prediction sets on average for both test datasets compared to Base, APS, and RAPS, while guaranteeing the marginal coverage of true labels as $1-\delta$ in the produced prediction sets. Note that regardless of how hyperparameters are chosen in RAPS (even if RAPS is optimized to have small sets), ECP outperforms RAPS in terms of efficiency by producing smaller and more precise prediction sets. 
Moreover, the variance (fluctuations) in marginal coverage is higher when testing on ImageNet-V2 due to having less data and different distribution compared to training data. Nevertheless, ECP is also able to produce smaller sets and still guarantee marginal coverage when tested on ImageNet-V2 if holdout set is selected from the new distribution.
We observe a similar behaviour when $\delta=0.05$. We report the results for $\delta=0.05$ in Appendix~\ref{size_append} to highlight that our proposed method is compatible with higher guarantees of marginal coverage and its ability to produce smaller sets compared to SoTA methods. Note that in both test sets, we use significantly smaller holdout set to quantify the quantile compared to the results already reported in RAPS method~\citep{cp_raps}. This shows the strength of ECP that even with limited holdout data, ECP outperforms SoTA methods in terms of efficiency, i.e. achieving smaller and more precise prediction sets. 

\begin{table}[t]
\centering
\caption{Efficiency on \textbf{ImageNet-V2}: The median-of-means for marginal coverage and prediction set size using different architectures and methods with $\pmb{\delta=0.1}$ over 10 trials.}
\scalebox{0.85}{
\begin{tabular}{l c cccc ccc>{\bfseries}c}
    \toprule
\multirow{0}{*}{\textbf{Model}} 
        & \multicolumn{1}{c}{\textbf{Performance}} & \multicolumn{4}{c}{\textbf{Mean Marginal Coverage}} &
        \multicolumn{4}{c}{\textbf{Mean Prediction Set Size}} \\
    \cmidrule(lr){2-2} \cmidrule(lr){3-6} \cmidrule(lr){7-10}
        & Accuracy & Base & APS & RAPS & ECP & Base & APS & RAPS & ECP (Ours) \\
    \midrule
ResNeXT101 
        & 67.51\%  & 0.891 & 0.901 & 0.900 & 0.900 & 35.32 & 54.67 & 5.56 & 4.88 \\
    \addlinespace
ResNet152
        & 66.98\% & 0.895 & 0.900 & 0.900 & 0.900 & 25.11 & 29.67 & 5.52 & 4.28 \\
    \addlinespace
ResNet101
        & 65.54\% & 0.893 & 0.900 & 0.899 & 0.899 & 29.06 & 33.34 & 8.66 & 5.65 \\
    \addlinespace
ResNet50
        & 63.20\%  & 0.884 & 0.899 & 0.901 & 0.900 & 31.46 & 33.53 & 9.34 & 6.68 \\
    \addlinespace
ResNet18
        & 57.29\% & 0.896 & 0.900 & 0.900 & 0.901 & 38.85 & 40.03 & 17.71 & 11.68 \\
    \addlinespace
DenseNet161
        & 65.13\% & 0.891 & 0.900 & 0.900 & 0.900 & 28.18 & 32.14 & 6.42 & 5.72 \\
    \addlinespace
VGG16
        & 58.81\% & 0.881 & 0.899 & 0.898 & 0.900 & 31.56 & 32.91 & 13.85 & 9.74 \\
    \addlinespace
ShuffleNet
        & 56.04\% & 0.885 & 0.901 & 0.900 & 0.899 & 66.40 & 75.84 & 24.99 & 17.90 \\
    \addlinespace
Inception
        & 57.59\%  & 0.891 & 0.900 & 0.900 & 0.900 & 132.75 & 158.49 & 16.77 & 15.09 \\
    % \addlinespace

    \bottomrule
\end{tabular}  }
\label{perform_imagenetv2}
\end{table}

\subsection{Experiment 2: Coverage in Size-stratified Prediction Sets}
\label{cov_size_strat_exp}
In this experiment, we stratify the size of the prediction set produced by our method and the other three SoTA methods into different ranges \{$0-1$, $2-3$, $4-6$, $7-10$, $11-100$, $101-1000$\} to evaluate the coverage of prediction sets in each size range. We use ResNet-152 on ImageNet-Val with the coverage error level $\delta=0.1$ and different choices of hyperparameters $\lambda$ for RAPS. 

\vspace{1mm}
\noindent \textbf{Observations.} In Table~\ref{size_strat_result}, we observe that APS covers true labels in prediction sets of sizes $101-1000$ with a high probability around $97\%$ that is much higher than the desired coverage 90\%, i.e., APS aims to guarantee marginal coverage by undercovering easy examples (smaller set sizes) and overcovering difficult examples (larger set sizes) which is not an acceptable behavior since APS is far from the \emph{exact conditional coverage}. Note that the exact conditional coverage is defined for a specific test data point (conditionality) as $\mathcal{P}(y_{val} \in \mathcal{C}(X,\widehat{q})\ |\ X=x_{val})=1-\delta$. 
For RAPS, when $\lambda$ is small, RAPS produces larger sets. However, when $\lambda$ is high, RAPS produces smaller sets (at most with the size $k_{reg}$), but with less stability in conditional coverage since as shown in~\tableref{size_strat_result}, with high $\lambda$, the coverage of RAPS for larger set sizes are farther from exact conditional coverage, i.e. $1-\delta$, compared to ECP. Thus, ECP is more likely to have or to be near exact conditional coverage rather than RAPS when produces smaller sets with high $\lambda$ although ECP always produces smaller sets rather than RAPS even in these cases. Note that RAPS with light regularization (lower $\lambda$) is stable and more balanced compared to when $\lambda$ is high.

%%%% DISCUSSION ON COUNTS MAYBE NEEDED HERE AFTER ENTERING THE VALUES OF TABLES.

\begin{table}[t]
\centering
\caption{The count and average coverage of images stratified by their prediction set sizes on ImageNet-Val using ResNet-152 with $\delta=0.1$ and different $\lambda$ to evaluate APS and RAPS.}
\scalebox{0.80}{
\begin{tabular}{l cc cc cc cc cc cc}
    \toprule
\multirow{0}{*}{\textbf{Set Size}} 
        & \multicolumn{2}{c}{\textbf{ECP (Ours)}} & \multicolumn{2}{c}{\textbf{APS ($\pmb{\lambda=0}$)}} &
        \multicolumn{2}{c}{\textbf{$\pmb{\lambda=0.001}$}} & \multicolumn{2}{c}{\textbf{$\pmb{\lambda=0.01}$}} &
        \multicolumn{2}{c}{\textbf{$\pmb{\lambda=0.1}$}} & \multicolumn{2}{c}{\textbf{$\pmb{\lambda=1}$}} \\
    \cmidrule(lr){2-3} \cmidrule(lr){4-5} \cmidrule(lr){6-7} \cmidrule(lr){8-9} \cmidrule(lr){10-11} \cmidrule(lr){12-13}
        & cnt & cvg & cnt & cvg & cnt & cvg & cnt & cvg & cnt & cvg & cnt & cvg \\
    \midrule
0 to 1
        & 18322  & 0.932 & 20149 & 0.880 & 20177 & 0.879 & 19416 & 0.894 & 18422 & 0.910  & 17521 & 0.927\\
    \addlinespace
2 to 3
        & 13981 & 0.874 & 6524 & 0.910 & 6418 &  0.913 & 6643 & 0.923 & 6764 & 0.928 & 6824 & 0.941\\
    \addlinespace
4 to 6
        & 2639 & 0.833 & 2196 & 0.913 & 2313 & 0.922 & 2885 & 0.929 & 6540 & 0.904 & 10655 & 0.833\\
    \addlinespace
7 to 10
        & 58 &  0.880 &  1242  &  0.927  & 1313 &  0.923 & 2259 & 0.915 & 3170 & 0.758 & 0 & -\\
    \addlinespace
11 to 100
        & 0 &  -  &  3873  &  0.951  & 4583 &  0.938 & 3797 & 0.875 & 104 & 0.471 & 0 & -\\
    \addlinespace
101 to 1000
        & 0 &  -  &  1016  &  0.972  & 196 &  0.903 & 0 & - & 0 & - & 0 & -\\
    % \addlinespace

    \bottomrule
\end{tabular}  }
\label{size_strat_result}
\end{table}

\subsection{Experiment 3: Adaptivity in Prediction Sets}
\label{adapt_set_exp}

\begin{figure}[t] 
\centering
\subfigure[ImageNet-Val]{
    \includegraphics[width=0.47\linewidth]{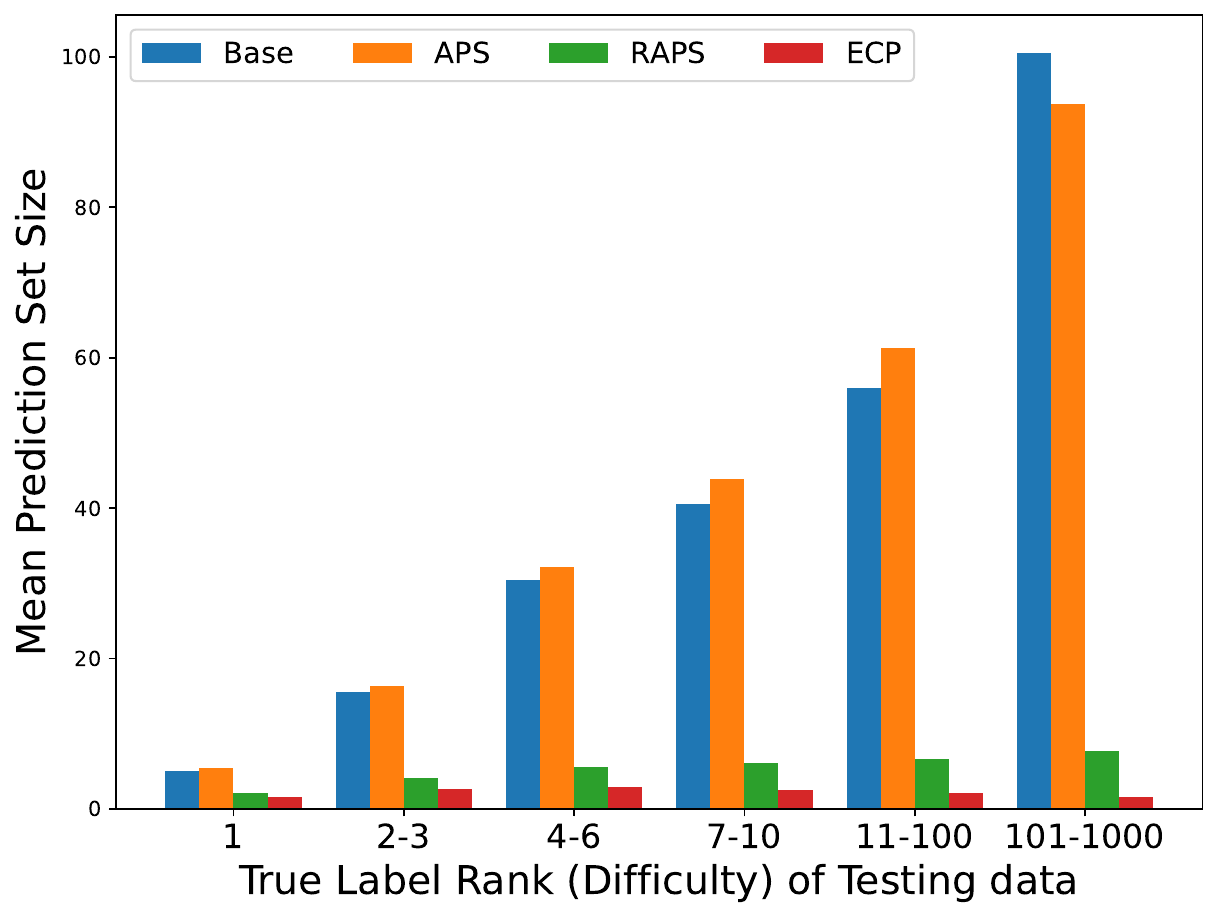}%
    \label{adapt_val}
    }
~
\subfigure[ImageNet-V2]{
    \includegraphics[width=0.47\linewidth]{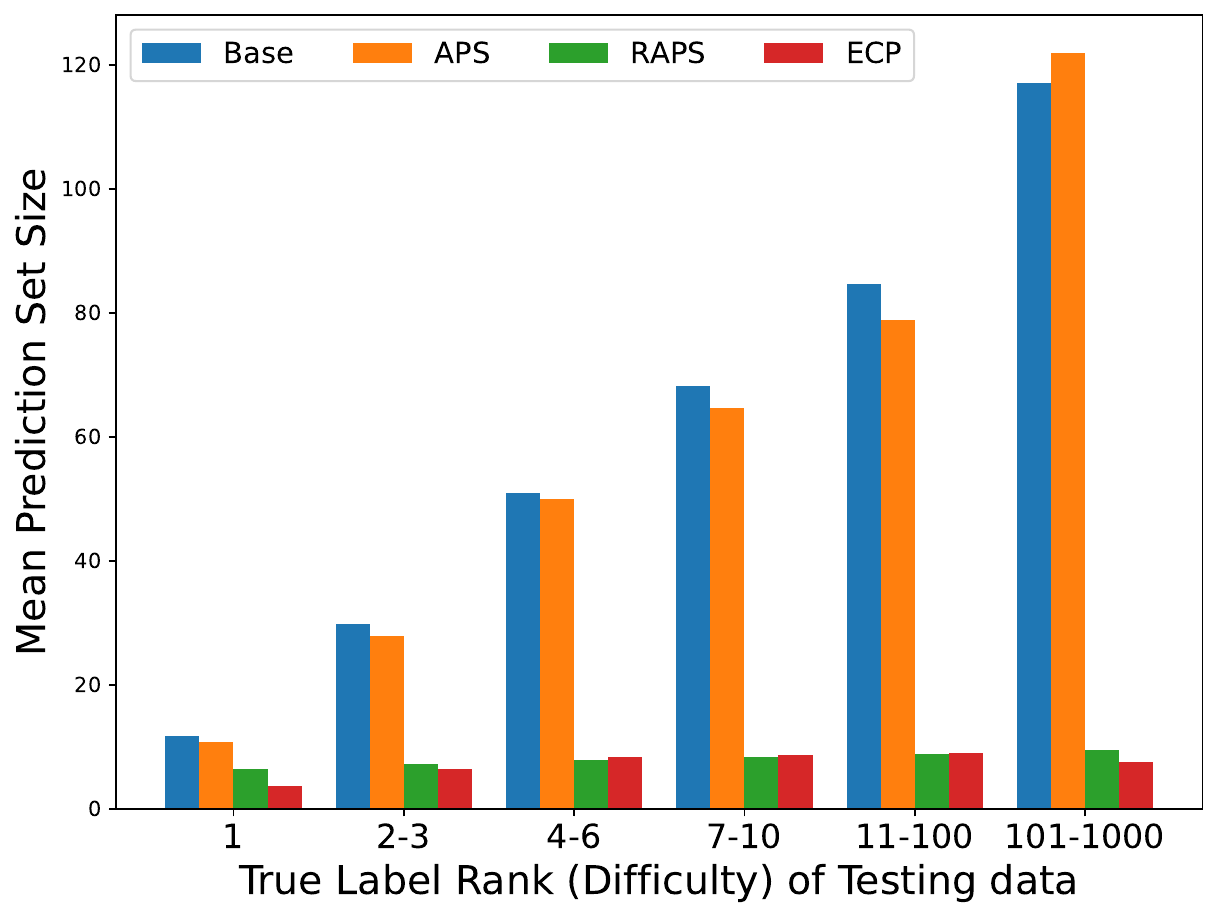}%
    \label{adapt_v2}
    }
\caption{Adaptivity in ECP compared to SoTA methods based on difficulty level for both test datasets using ResNet-152 with $\delta=0.1$. For RAPS, we used $k_{reg}=5$ and $\lambda=0.1$.}
\label{adapt_ecp_diff}
\end{figure}

In this section, we address the adaptivity criterion to evaluate our method compared to the other methods based on difficulty level of test images. The difficulty of an image is determined and represented by the rank of the true label's predictive probability in the descending-sorted set of predictive probabilities associated with the target labels. 
\tableref{diff_result} shows the average marginal coverage and set size for the methods based on different ranges of difficulty level in test data. We stratify the image difficulty levels into different ranges \{$1$, $2-3$, $4-6$, $7-10$, $11-100$, $101-1000$\} to evaluate and compare the adaptivity of the methods in each difficulty range. 
Moreover, \figureref{adapt_ecp_diff} shows the average set size for the methods based on different difficulty ranges for both ImageNet-Val (\figureref{adapt_val}) and ImageNet-V2 (\figureref{adapt_v2}).

\vspace{1mm}
\noindent \textbf{Observations.} The results show that ECP produces smaller prediction sets for easy images (low difficulty levels) compared to the difficult ones. 
We observe that ECP maintain its adaptivity while producing smaller set sizes on average rather than other SoTA methods when testing with both datasets. 
% By applying ECP, there may be a decrease in set sizes for images with high difficulty levels since unlike APS and RAPS, when increasing the difficulty level, the standard deviation (std) of set sizes is not continuously increased and instead, it may be decreased. 
Because of trade-off between adaptiveness and set size, ECP aims to produce the smallest possible sets for difficult images (even empty sets) where there is no significant or useful coverage in prediction sets as shown in~\tableref{diff_result}. We can see the poor coverage for both ECP and RAPS methods in predicting difficult images due to producing small sets compared to APS method. Thus, ECP uses this phenomenon to further reduce the set sizes for difficult images, even with producing more empty sets, while their sizes are significantly larger than easy images to preserve adaptivity. Although ECP and RAPS (with high $\lambda$) both cover difficult images less than $1-\delta$ to produce smaller sets, ECP compensates it by carefully having slightly larger sets for easy images which leads to higher coverage compared to RAPS to guarantee $1-\delta$ marginal coverage on average. Note that this increase in set size for easy images is adequately slight due to the higher number of easy images rather than difficult ones.

\begin{table}[t]
\centering
\caption{The average marginal coverage and set size of images stratified by their difficulty levels, i.e. the rank of true label's predictive probability, on ImageNet-Val using ResNet-152 with $\delta=0.1$ and different $\lambda$ to evaluate APS and RAPS.}
\scalebox{0.80}{
\begin{tabular}{l ccc c cc cc cc cc}
    \toprule
\multirow{0}{*}{\textbf{Difficulty}} & \multicolumn{3}{c}{\textbf{ECP (Ours)}} 
            & \multirow{1}{*}{\textbf{APS/RAPS}} & \multicolumn{2}{c}{\textbf{APS ($\pmb{\lambda=0}$)}} &
         \multicolumn{2}{c}{\textbf{$\pmb{\lambda=0.01}$}} &
        \multicolumn{2}{c}{\textbf{$\pmb{\lambda=0.1}$}} & \multicolumn{2}{c}{\textbf{$\pmb{\lambda=1}$}} \\
    \cmidrule(lr){2-4} \cmidrule(lr){5-5} \cmidrule(lr){6-7} \cmidrule(lr){8-9} \cmidrule(lr){10-11} \cmidrule(lr){12-13}
        & cnt & cvg & sz & cnt & cvg & sz & cvg & sz & cvg & sz & cvg & sz \\
    \midrule
1
        & 27336 & 0.999 & 1.47 & 27428 & 0.951 & 5.44 & 0.959 & 2.60 & 0.970 & 2.07 & 0.977 & 2.02\\
    \addlinespace
2 to 3
        & 4573 & 0.817 & 2.56 & 4524 & 0.783 & 16.19 & 0.810 & 6.27 & 0.830 & 4.04 & 0.866 & 3.64\\
    \addlinespace
4 to 6
        & 1239 & 0.203 & 2.84 & 1229 & 0.695 & 32.32 & 0.726 & 9.97 & 0.727 & 5.49 & 0.659 & 4.57\\
    \addlinespace
7 to 10
        & 621 & 0.005 & 2.64 & 598 &  0.640 & 41.53 & 0.624 & 11.97 & 0.247 & 5.89 & 0 & 4.61\\
    \addlinespace
11 to 100
        & 1092 & 0 & 2.32 & 1085 &  0.562 & 60.02 & 0.281 & 15.52 & 0 & 6.65 & 0 & 4.88\\
    \addlinespace
101 to 1000
        & 139 & 0 & 2.00 & 136 &  0.220 & 101.42 & 0 & 20.95 & 0 & 7.33 & 0 & 5.02\\
    % \addlinespace

    \bottomrule
\end{tabular}  }
\label{diff_result}
\end{table}

\subsection{Experiment 4: Adaptivity in Conditional Coverage}
\label{adapt_cov_exp}

\begin{table}[t]
\centering
\caption{The quality of prediction sets of ImageNet-Val data based on the average of size-adaptivity trade-off (SAT) score with respect to average set size and SSCV in different methods and model architectures with $\delta=0.1$.}
\scalebox{1.0}{
\begin{tabular}{l ccc ccc >{\bfseries}cc>{\bfseries}c}
    \toprule
\multirow{0}{*}{\textbf{Model}} 
        & \multicolumn{3}{c}{\textbf{APS}} & \multicolumn{3}{c}{\textbf{RAPS}} & \multicolumn{3}{c}{\textbf{ECP (Ours)}} \\
    \cmidrule(lr){2-4} \cmidrule(lr){5-7} \cmidrule(lr){8-10}
        & Size & SSCV & SAT & Size & SSCV & SAT & Size & SSCV & SAT \\
    \midrule
ResNeXT101 
        & 19.21 & 0.088 & 0.045 & 4.43 & 0.029 & 0.212 & 1.68 & 0.061 & 0.551 \\
    \addlinespace
ResNet152
        & 10.85 & 0.072 & 0.082 & 3.93 & 0.025 & 0.239 & 1.77 & 0.077 & 0.517 \\
    \addlinespace
ResNet101
        & 11.01  & 0.071 & 0.081 & 4.21 & 0.024 & 0.224 & 1.85 & 0.076 & 0.481 \\
    \addlinespace
ResNet50
        & 12.59 &  0.069 & 0.071 & 4.54 & 0.029 & 0.207 & 2.03 & 0.080 & 0.411 \\
    \addlinespace
ResNet18
        & 16.44 & 0.050 & 0.056 & 5.43 & 0.002 & 0.183 & 3.68 & 0.075 & 0.236 \\
    \addlinespace
DenseNet161
        & 11.82 & 0.077 & 0.075 & 4.17 & 0.020 & 0.227 & 1.85 & 0.070 & 0.470 \\
    \addlinespace
VGG16
        & 14.06 & 0.049 & 0.065 & 8.92 & 0.016 & 0.107 & 2.97 & 0.067 & 0.292 \\
    \addlinespace
ShuffleNet
        & 31.24 & 0.059 & 0.029 & 6.31 &  0.001 & 0.158 & 4.01 & 0.128 & 0.227 \\
    \addlinespace
Inception
        & 87.87 & 0.084 & 0.010 & 5.87 & 0.002 & 0.169 & 4.15 & 0.132 & 0.241 \\
    % \addlinespace

    \bottomrule
\end{tabular}  }
\label{sat_result}
\end{table}

In this experiment, we evaluate ECP compared to other methods based on the adaptivity in conditional coverage which is considered a promising approach to assess adaptiveness in image classifiers~\citep{cp_raps}. 
% As we previously mentioned in~\sectionref{cov_size_strat_exp}, 
A set predictor function $\mathcal{C}:\mathbb{R}^d\rightarrow 2^{\mathcal{Y}}$ satisfies exact conditional coverage for each $x_{val}\in \mathcal{X}_{val}$ if $\mathcal{P}(y\in \mathcal{C}(X)\ |\ X=x_{val})=1-\delta$. Practically, this exact conditional coverage is infeasible in distribution-free setting, i.e. guaranteeing exact conditional coverage requires distributional assumptions on input data~\citep{cp_conditional}, but it can be satisfied approximately. Furthermore, exact conditional coverage requires a perfect classifier (i.e., trained on infinite data) with high accuracy, and cannot be achievable in real world due to limited and realistic sample sizes during training. Alternatively, RAPS followed the notion of local conditional coverage~\citep{local_cp} where the coverage in a neighborhood of each unseen data point is considered and weighted based on the size of the neighborhood. Therefore, if exact conditional coverage holds, then it holds for size-stratified prediction sets~\citep{cp_raps}. Formally, if $\mathcal{P}(y\in \mathcal{C}(X)\ |\ X=x_{val})=1-\delta$, then $\mathcal{P}(y\in \mathcal{C}(X)\ |\ |\mathcal{C}(X)| \in \mathcal{S})=1-\delta$ for any $\mathcal{S} \subset \{0,1,2,...,K\}$ where $K=|\mathcal{Y}|$ is the number of labels and $\mathcal{S}$ is a subset of different sizes for prediction sets. In practice, a relaxation is considered for coverage of size-stratified sets as $\mathcal{P}(y\in \mathcal{C}(X)\ |\ |\mathcal{C}(X)| \in \mathcal{S})\geq 1-\delta$ for any $\mathcal{S} \subset \{0,1,2,...,K\}$.

A measure of adaptivity in prediction sets is introduced in~\citep{cp_raps} as \emph{Size-stratified Coverage Violation (SSCV)}, which is based on the violations from size-stratified conditional coverage. Consider $\mathcal{S}_i \subset \{0,1,2,...,K\}$ are disjoint set-size ranges such that $\bigcup_{i=1}^{i=s} \mathcal{S}_i=\{0,1,2,...,K\}$, and $\mathcal{I}_i$ is the set of data indices $j$ that their prediction set sizes are in set-size range $\mathcal{S}_i$ such that $\mathcal{I}_i=\{j: |\mathcal{C}(X_j)|\in \mathcal{S}_i\}$. Then, the metric SSCV for the set-valued predictor $\mathcal{C}$ and set-size ranges $\{\mathcal{S}_i\}_{i=1}^{i=s}$ is defined as,   
\begin{equation}
SSCV(\mathcal{C},\{\mathcal{S}_i\}_{i=1}^{i=s})=\sup_i\Bigg|\frac{\big|\{j:y_j\in\mathcal{C}(X_j), j\in \mathcal{I}_i\}\big|}{|\mathcal{I}_i|}-(1-\delta)\Bigg|\ , 
\label{sscv_eq}
\end{equation}
where $0\leq SSCV(\mathcal{C},\{\mathcal{S}_i\}_{i=1}^{i=s})<1$ is the maximum amount of deviation from the exact conditional coverage of $1-\delta$ (i.e. worst case) in size-stratified sets $\{\mathcal{S}_i\}_{i=1}^{i=s}$ of prediction sets $\mathcal{C}(X)$.
Note that in case of exact conditional coverage, the worst violation will be zero and $SSCV(\mathcal{C},\{\mathcal{S}_i\}_{i=1}^{i=s})=0$. Thus, the lower SSCV represents higher adaptivity in prediction sets. RAPS is optimized to have the most adaptivity in coverage, i.e. the least SSCV, by choosing the optimal hyperparameter $\lambda^*$ such that $\lambda^*=\underset{\lambda}{\arg\min}\ SSCV(\mathcal{C},\{\mathcal{S}_i\}_{i=1}^{i=s})$.   

We also propose our measure of quality in prediction sets incorporating both the coverage adaptivity and the average set size simultaneously to achieve a better judgement on the quality of prediction sets. We introduce a metric called \emph{Size-Adaptivity Trade-off (SAT)} that increases when the average set size (reported in~\tableref{perform_imagenetval,perform_imagenetv2}) is decreased and/or the coverage adaptivity is increased (SSCV is decreased). Therefore, SAT score is inversely proportional to both the average set size $1\leq \mu(\mathcal{C})\leq K$ and the metric of adaptivity $SSCV(\mathcal{C},\{\mathcal{S}_i\}_{i=1}^{i=s})$ computed in~\equationref{sscv_eq}.  
We define SAT score for the set-valued predictor $\mathcal{C}$ and set-size ranges $\{\mathcal{S}_i\}_{i=1}^{i=s}$ as,
\begin{equation}
    SAT(\mathcal{C},\{\mathcal{S}_i\}_{i=1}^{i=s}) = \frac{1-SSCV(\mathcal{C},\{\mathcal{S}_i\}_{i=1}^{i=s})}{\mu(\mathcal{C})} \qquad \text{s.t.} \qquad \mu(\mathcal{C})=\frac{\displaystyle\sum_{x\in \mathcal{X}_{val}}\big|\mathcal{C}(X=x)\big|}{\displaystyle\sum_{x\in \mathcal{X}_{val}} \mathbbm{1}_{\{|\mathcal{C}(X=x)|>0\}}}\ ,
\label{sat_eq}
\end{equation}
where $\mathbbm{1}$ denotes the indicator function over non-empty prediction sets, and $0<SAT(.)\leq 1$ represents the quality of set-valued predictor $\mathcal{C}$.
Note that as empty prediction sets have no practical information regarding the predictive uncertainty during deployment and a large number of empty sets can significantly reduce the average set size but without efficiency, we skip the empty prediction sets when computing $\mu(\mathcal{C})$ in~\equationref{sat_eq}. In case of using a perfect classifier that is trained on infinite data and perfectly fit the true label for given input data, $SSCV(\mathcal{C},\{\mathcal{S}_i\}_{i=1}^{i=s})=0$ due to the exact conditional coverage, and the model produces prediction sets with the fixed size of $1$ only containing the true label, i.e., $\mu(\mathcal{C})=1$; thus, the maximum SAT score is achieved, i.e. $SAT(\mathcal{C},\{\mathcal{S}_i\}_{i=1}^{i=s})=1$. 
In our experiments, we partition the produced set sizes into $0-1$, $2-3$, $4-10$, $11-100$, and $101-1000$ ranges. To apply RAPS method, we choose the optimal $\lambda$ that yields to the smallest SSCV score out of the set of $\lambda \in \{1e-5, 1e-4, 8e-4, 1e-3, 15e-4, 2e-3, 1e-2, 0.1, 1\}$. 
In~\tableref{sat_result}, we demonstrate the average SAT scores for ECP (the boldfaced columns) compared to other methods along with their corresponding SSCV value and set sizes considering different models. We also conducted further experiments to report and compare the adaptivity and the quality of prediction sets for ECP and Least Ambiguous Set-valued classifier (LAS) method in~\appendixref{lac_comp_append}.    

\vspace{1mm}
\noindent \textbf{Observations.} We observe that ECP outperforms SoTA methods with SAT scores that are consistently higher than the other methods indicating that the prediction sets produced by ECP have noticeably higher quality rather than APS and RAPS. Practically speaking, higher SAT in ECP shows better trade-off between efficiency (set size) and coverage adaptivity as two conflicting criteria rather than SoTA methods. 
Although APS is an adaptive method with small SSCV scores similar to ECP, its SAT scores are significantly smaller than other methods due to high inefficiency (very large sets) and it also does not satisfy conditional coverage shown in Experiment~\ref{cov_size_strat_exp} and~\tableref{size_strat_result}. 

When applying RAPS, we find the optimal $\lambda$ where SSCV score for prediction sets is minimum. In practice, the optimal $\lambda$ is usually small since RAPS produces larger sets when $\lambda$ is smaller (light regularization) with near exact coverage, i.e. small violations from $1-\delta$ coverage, conditional on size-stratified sets (see~\tableref{diff_result}). This further optimization for addressing only one criterion (adaptivity) can also add computational burden compared to ECP method that is stable and free from hyperparameters. 
However, smaller SSCV scores in RAPS only imply on better adaptiveness which is not adequate for conclusion on the quality of prediction sets since the set size as a practical criterion is ignored. Therefore, unlike ECP, RAPS can be either optimized for smaller sets or more coverage adaptivity at a time. Although it gives us a flexibility to choose the required behavior, the desired prediction sets are the ones that consider all the three criteria, i.e. coverage, efficiency, and adaptivity, simultaneously. In case of conflicting criteria, a trade-off is required to produce prediction sets which is considered in ECP rather than SoTA methods, particularly the trade-off between set size and adaptivity. 
As shown in~\tableref{sat_result}, we can see that although RAPS shows more adaptiveness with slightly smaller SSCV compared to ECP, RAPS significantly sacrifices the efficiency and shows larger sets. 
Thus, SAT scores for ECP are considerably higher rather than APS and RAPS by considering both adaptiveness and efficiency of prediction sets, simultaneously. This implies that the prediction sets produced by ECP is more practical having higher quality rather than the other methods.

\section{Conclusion}
\label{conc}
In this paper, we have addressed the problem of model uncertainty in deep classifiers using
% it used to be image classifiers here before CR version
conformal prediction. We used the model (epistemic) uncertainty derived from EDL as our heuristic notion of uncertainty along with evidential properties such as uncertainty surprisal and expected utility associated with the target labels, to propose a novel non-conformity score function. We applied our non-conformity score function to introduce evidential CP (ECP) and produce efficient and adaptive prediction sets while maintaining the coverage of true labels. Our extensive experimental evaluations showed that ECP could outperform the SoTA CP methods in terms of efficiency (small set size) and adaptivity with no additional optimization process or computational burden. 

% While our research has made notable contributions, there are still opportunities for further exploration. Future work should focus on addressing challenges such as high-dimensional data, imbalanced datasets, and incorporating domain knowledge into uncertainty quantification in conformal prediction. Additionally, investigating interpretability and explainability of uncertainty measures can provide actionable insights. We encourage continued research to foster the development of more reliable and accurate uncertainty quantification methods within the conformal prediction framework.

% In summary, this paper advances model uncertainty quantification in conformal prediction, offering an improved technique to compare CP-based methods with other state-of-the-art methods. 

\acks{We would like to express our sincere gratitude to the anonymous reviewers for their valuable comments, which greatly improved the quality of this manuscript. This research is supported by Natural Sciences and Engineering Research Council of Canada (NSERC) Discovery Grant RGPIN-2016-06062.}

\bibliography{copa2024}

\begin{thebibliography}{30}
\providecommand{\natexlab}[1]{#1}
\providecommand{\url}[1]{\texttt{#1}}
\expandafter\ifx\csname urlstyle\endcsname\relax
  \providecommand{\doi}[1]{doi: #1}\else
  \providecommand{\doi}{doi: \begingroup \urlstyle{rm}\Url}\fi

\bibitem[Amini et~al.(2020)Amini, Schwarting, Soleimany, and Rus]{edl_reg}
Alexander Amini, Wilko Schwarting, Ava Soleimany, and Daniela Rus.
\newblock Deep evidential regression.
\newblock \emph{Advances in Neural Information Processing Systems}, 33:\penalty0 14927--14937, 2020.

\bibitem[Angelopoulos et~al.(2020)Angelopoulos, Bates, Malik, and Jordan]{cp_raps}
Anastasios Angelopoulos, Stephen Bates, Jitendra Malik, and Michael~I Jordan.
\newblock Uncertainty sets for image classifiers using conformal prediction.
\newblock \emph{arXiv preprint arXiv:2009.14193}, 2020.

\bibitem[Angelopoulos et~al.(2023)Angelopoulos, Bates, et~al.]{cp_book}
Anastasios~N Angelopoulos, Stephen Bates, et~al.
\newblock Conformal prediction: A gentle introduction.
\newblock \emph{Foundations and Trends{\textregistered} in Machine Learning}, 16\penalty0 (4):\penalty0 494--591, 2023.

\bibitem[Dempster(1968)]{dst}
Arthur~P Dempster.
\newblock A generalization of bayesian inference.
\newblock \emph{Journal of the Royal Statistical Society: Series B (Methodological)}, 30\penalty0 (2):\penalty0 205--232, 1968.

\bibitem[Gal and Ghahramani(2016)]{dropout}
Yarin Gal and Zoubin Ghahramani.
\newblock Dropout as a bayesian approximation: Representing model uncertainty in deep learning.
\newblock In \emph{international conference on machine learning}, pages 1050--1059. PMLR, 2016.

\bibitem[Guo et~al.(2017)Guo, Pleiss, Sun, and Weinberger]{calib1}
Chuan Guo, Geoff Pleiss, Yu~Sun, and Kilian~Q Weinberger.
\newblock On calibration of modern neural networks.
\newblock In \emph{International Conference on Machine Learning}, pages 1321--1330. PMLR, 2017.

\bibitem[He et~al.(2016)He, Zhang, Ren, and Sun]{resnet}
Kaiming He, Xiangyu Zhang, Shaoqing Ren, and Jian Sun.
\newblock Deep residual learning for image recognition.
\newblock In \emph{Proceedings of the IEEE conference on computer vision and pattern recognition}, pages 770--778, 2016.

\bibitem[Huang et~al.(2017)Huang, Liu, van~der Maaten, and Weinberger]{densenet}
Gao Huang, Zhuang Liu, Laurens van~der Maaten, and Kilian~Q. Weinberger.
\newblock Densely connected convolutional networks.
\newblock In \emph{Proceedings of the IEEE Conference on Computer Vision and Pattern Recognition (CVPR)}, July 2017.

\bibitem[J{\o}sang(2016)]{sl}
Audun J{\o}sang.
\newblock Subjective logic, 2016.

\bibitem[Karimi and Samavi(2023)]{uq_cp}
Hamed Karimi and Reza Samavi.
\newblock Quantifying deep learning model uncertainty in conformal prediction.
\newblock In \emph{Proceedings of the AAAI Symposium Series}, volume~1, pages 142--148, 2023.

\bibitem[Kotz et~al.(2019)Kotz, Balakrishnan, and Johnson]{beta_func}
Samuel Kotz, Narayanaswamy Balakrishnan, and Norman~L Johnson.
\newblock \emph{Continuous multivariate distributions, Volume 1: Models and applications}, volume 334.
\newblock John Wiley \& Sons, 2019.

\bibitem[Lakshminarayanan et~al.(2016)Lakshminarayanan, Pritzel, and Blundell]{ensemble}
Balaji Lakshminarayanan, Alexander Pritzel, and Charles Blundell.
\newblock Simple and scalable predictive uncertainty estimation using deep ensembles.
\newblock \emph{arXiv preprint arXiv:1612.01474}, 2016.

\bibitem[Ma et~al.(2018)Ma, Zhang, Zheng, and Sun]{shufflenet}
Ningning Ma, Xiangyu Zhang, Hai-Tao Zheng, and Jian Sun.
\newblock Shufflenet v2: Practical guidelines for efficient cnn architecture design.
\newblock In \emph{Proceedings of the European conference on computer vision (ECCV)}, pages 116--131, 2018.

\bibitem[Nixon et~al.(2019)Nixon, Dusenberry, Zhang, Jerfel, and Tran]{calib2}
Jeremy Nixon, Michael~W Dusenberry, Linchuan Zhang, Ghassen Jerfel, and Dustin Tran.
\newblock Measuring calibration in deep learning.
\newblock In \emph{CVPR Workshops}, volume~2, 2019.

\bibitem[Papadopoulos et~al.(2002)Papadopoulos, Proedrou, Vovk, and Gammerman]{inductive_cp}
Harris Papadopoulos, Kostas Proedrou, Volodya Vovk, and Alex Gammerman.
\newblock Inductive confidence machines for regression.
\newblock In \emph{Machine Learning: ECML 2002: 13th European Conference on Machine Learning Helsinki, Finland, August 19--23, 2002 Proceedings 13}, pages 345--356. Springer, 2002.

\bibitem[Pearce et~al.(2018)Pearce, Brintrup, Zaki, and Neely]{non_cp1}
Tim Pearce, Alexandra Brintrup, Mohamed Zaki, and Andy Neely.
\newblock High-quality prediction intervals for deep learning: A distribution-free, ensembled approach.
\newblock In \emph{International conference on machine learning}, pages 4075--4084. PMLR, 2018.

\bibitem[Platt et~al.(1999)]{platt}
John Platt et~al.
\newblock Probabilistic outputs for support vector machines and comparisons to regularized likelihood methods.
\newblock \emph{Advances in large margin classifiers}, 10\penalty0 (3):\penalty0 61--74, 1999.

\bibitem[Romano et~al.(2020)Romano, Sesia, and Candes]{cp_aps}
Yaniv Romano, Matteo Sesia, and Emmanuel Candes.
\newblock Classification with valid and adaptive coverage.
\newblock \emph{Advances in Neural Information Processing Systems}, 33:\penalty0 3581--3591, 2020.

\bibitem[Sadinle et~al.(2019)Sadinle, Lei, and Wasserman]{cp_lac}
Mauricio Sadinle, Jing Lei, and Larry Wasserman.
\newblock Least ambiguous set-valued classifiers with bounded error levels.
\newblock \emph{Journal of the American Statistical Association}, 114\penalty0 (525):\penalty0 223--234, 2019.

\bibitem[Sensoy et~al.(2018)Sensoy, Kaplan, and Kandemir]{evid1}
Murat Sensoy, Lance Kaplan, and Melih Kandemir.
\newblock Evidential deep learning to quantify classification uncertainty.
\newblock \emph{Advances in neural information processing systems}, 31, 2018.

\bibitem[Shafer(1976)]{evid_math}
Glenn Shafer.
\newblock A mathematical theory of evidence, 1976.

\bibitem[Simonyan and Zisserman(2014)]{vgg}
Karen Simonyan and Andrew Zisserman.
\newblock Very deep convolutional networks for large-scale image recognition.
\newblock \emph{arXiv preprint arXiv:1409.1556}, 2014.

\bibitem[Szegedy et~al.(2016)Szegedy, Vanhoucke, Ioffe, Shlens, and Wojna]{inception}
Christian Szegedy, Vincent Vanhoucke, Sergey Ioffe, Jon Shlens, and Zbigniew Wojna.
\newblock Rethinking the inception architecture for computer vision.
\newblock In \emph{Proceedings of the IEEE conference on computer vision and pattern recognition}, pages 2818--2826, 2016.

\bibitem[Tibshirani et~al.(2019)Tibshirani, Foygel~Barber, Candes, and Ramdas]{local_cp}
Ryan~J Tibshirani, Rina Foygel~Barber, Emmanuel Candes, and Aaditya Ramdas.
\newblock Conformal prediction under covariate shift.
\newblock \emph{Advances in neural information processing systems}, 32, 2019.

\bibitem[Tonkens et~al.(2023)Tonkens, Sun, Yu, and Herbert]{robotics_cp}
Sander Tonkens, Sophia Sun, Rose Yu, and Sylvia Herbert.
\newblock Scalable safe long-horizon planning in dynamic environments leveraging conformal prediction and temporal correlations.
\newblock In \emph{Long-Term Human Motion Prediction Workshop, International Conference on Robotics and Automation}, 2023.

\bibitem[Vazquez and Facelli(2022)]{medical_cp}
Janette Vazquez and Julio~C Facelli.
\newblock Conformal prediction in clinical medical sciences.
\newblock \emph{Journal of Healthcare Informatics Research}, 6\penalty0 (3):\penalty0 241--252, 2022.

\bibitem[Vovk(2012)]{cp_conditional}
Vladimir Vovk.
\newblock Conditional validity of inductive conformal predictors.
\newblock In \emph{Asian conference on machine learning}, pages 475--490. PMLR, 2012.

\bibitem[Vovk et~al.(2005)Vovk, Gammerman, and Shafer]{cp}
Vladimir Vovk, Alexander Gammerman, and Glenn Shafer.
\newblock Algorithmic learning in a random world, 2005.

\bibitem[Xie et~al.(2017)Xie, Girshick, Doll{\'a}r, Tu, and He]{resnext}
Saining Xie, Ross Girshick, Piotr Doll{\'a}r, Zhuowen Tu, and Kaiming He.
\newblock Aggregated residual transformations for deep neural networks.
\newblock In \emph{Proceedings of the IEEE conference on computer vision and pattern recognition}, pages 1492--1500, 2017.

\bibitem[Zhang et~al.(2018)Zhang, Wang, and Qiao]{non_cp2}
Chong Zhang, Wenbo Wang, and Xingye Qiao.
\newblock On reject and refine options in multicategory classification.
\newblock \emph{Journal of the American Statistical Association}, 113\penalty0 (522):\penalty0 730--745, 2018.

\end{thebibliography}

\newpage
\appendix

\section{Proof of~\theoremref{conf_cov_theorem}}
\label{proof_append}
\begin{proof} 
Following the belief and uncertainty quantification in EDL described in~\sectionref{edl_back}, we map the feature of Beta-distributed coverage to evidential setting to quantify the confidence and consequently, the uncertainty associated with the true label coverage. As Beta distribution is a Dirichlet distribution restricted by only two shape parameters (binomial version), we use evidential setting using a Beta distribution to form belief masses associated with each one of the parameters. Then, according to~\equationref{belief}, we can quantify the confidence (belief) of covering the true labels and the uncertainty associated with the coverage process. Formally, according to~\equationref{cov_dist}, the two parameters for the Beta-distributed coverage are $\alpha_1=n+1-l$ and $\alpha_2=l$ representing the strength of Beta distribution in covering and miscovering the true label, respectively. Note that as Beta distribution has only two parameters $\alpha_1$ and $\alpha_2$, we have $K=2$. Here, we have $\alpha_0=\sum_{i=1}^{K}\alpha_i=n+1$ as the summation of Beta parameters. The evidence associated with these parameters represent the amount of support for covering the true label denoted by $e_1$ or miscovering the true label denoted by $e_2$ and calculated as $e_1 = \alpha_1 - 1$ and $e_2 = \alpha_2 - 1$, respectively based on~\equationref{dirich_parameter}. Then, we can compute Beta parameters as $e_1 = n - l$ and $e_2 = l - 1$.
Now, we compute the belief $b_1$ as the amount of confidence $\gamma$ in covering the true label in the prediction sets on average corresponding to $e_1$ as,
\begin{equation}
\gamma = b_1 = \frac{e_1}{\alpha_0} = \frac{n-l}{n+1}=\frac{n-\lfloor (n+1)\delta \rfloor}{n+1}\ .
\label{}
\end{equation}
Also, the uncertainty associated with the coverage is defined according to~\equationref{belief} as,
\begin{equation}
    U_{\mathcal{C}}=\frac{K}{\alpha_0}=\frac{2}{n+1}\ .
\end{equation}

\end{proof}

\section{Efficiency of Prediction Sets When $\pmb{\delta=0.05}$}
\label{size_append}
In this section, we show the experimental results on the accuracy, average marginal coverage, and average prediction set size using different architectures and methods with $\delta=0.05$ on both test sets ImageNet-Val and ImageNet-V2 in~\tableref{eff_val_append,eff_v2_append}, respectively (same observations and conclusions as stated in~\sectionref{cov_size_exp}). 

\begin{table}[t]
\centering
\caption{Efficiency on \textbf{ImageNet-Val}: The median-of-means for marginal coverage and prediction set size using different architectures and methods with $\pmb{\delta=0.05}$ over 10 trials}
\scalebox{0.85}{
\begin{tabular}{l c cccc ccc>{\bfseries}c}
    \toprule
\multirow{0}{*}{\textbf{Model}} 
        & \multicolumn{1}{c}{\textbf{Performance}} & \multicolumn{4}{c}{\textbf{Mean Marginal Coverage}} &
        \multicolumn{4}{c}{\textbf{Mean Prediction Set Size}} \\
    \cmidrule(lr){2-2} \cmidrule(lr){3-6} \cmidrule(lr){7-10}
        & Accuracy & Base & APS & RAPS & ECP & Base & APS & RAPS & ECP (Ours) \\
    \midrule
ResNeXT101 
        & 79.31\%  & 0.937 & 0.950 & 0.950 & 0.950 & 36.09 & 45.73 & 4.14 & 3.11 \\
    \addlinespace
ResNet152
        & 78.31\%   & 0.941 & 0.950 & 0.950 & 0.950 &  19.30 & 22.22 & 4.43 & 3.24 \\
    \addlinespace
ResNet101
        & 77.37\%  & 0.945 & 0.951 & 0.950 & 0.951 & 19.62 & 23.31 & 4.87 & 3.83 \\
    \addlinespace
ResNet50
        & 76.13\%   &  0.942   &  0.950  &  0.950  & 0.950 &  21.81 & 24.37 & 7.22 & 4.44    \\
    \addlinespace
ResNet18
        & 69.76\%   &  0.944   &  0.951  &  0.949  & 0.950 & 28.85 & 31.44 & 13.65 & 8.85    \\
    \addlinespace
DenseNet161
        & 77.13\%   &  0.939   &  0.950  &  0.950  & 0.950 & 24.65 & 25.68 & 5.06 & 3.78 \\
    \addlinespace
VGG16
        & 71.59\%   &  0.945   &  0.950 &  0.901 & 0.950 &  25.31 & 28.69 & 11.10 & 6.73 \\
    \addlinespace
ShuffleNet
        & 69.36\%   &  0.941   &  0.951  &  0.950  & 0.949 &  62.23 & 71.90 & 16.87 & 11.65    \\
    \addlinespace
Inception
        & 69.54\%   & 0.946  &  0.951  &  0.950  & 0.950 & 138.43 & 171.52 & 19.32 & 10.96 \\
    % \addlinespace

    \bottomrule
\end{tabular}  }
\label{eff_val_append}
\end{table}

\begin{table}[t]
\centering
\caption{Efficiency on \textbf{ImageNet-V2}: The median-of-means for marginal coverage and prediction set size using different architectures and methods with $\pmb{\delta=0.05}$ over 10 trials}
\scalebox{0.85}{
\begin{tabular}{l c cccc ccc>{\bfseries}c}
    \toprule
\multirow{0}{*}{\textbf{Model}} 
        & \multicolumn{1}{c}{\textbf{Performance}} & \multicolumn{4}{c}{\textbf{Mean Marginal Coverage}} &
        \multicolumn{4}{c}{\textbf{Mean Prediction Set Size}} \\
    \cmidrule(lr){2-2} \cmidrule(lr){3-6} \cmidrule(lr){7-10}
        & Accuracy & Base & APS & RAPS & ECP & Base & APS & RAPS & ECP (Ours) \\
    \midrule
ResNeXT101 
        & 67.51\%  & 0.941 & 0.951 & 0.950 & 0.950 & 85.48 & 94.33 & 20.89 & 13.61 \\
    \addlinespace
ResNet152
        & 66.98\% & 0.943 & 0.949 & 0.950 & 0.951 & 52.90 & 57.07 & 26.30 & 11.36 \\
    \addlinespace
ResNet101
        & 65.54\% & 0.944 & 0.950 & 0.949 & 0.950 & 57.95 & 64.28 & 21.67 & 15.43 \\
    \addlinespace
ResNet50
        & 63.20\%  & 0.936 & 0.950 & 0.951 & 0.950 & 62.08 & 69.21 & 29.53 & 17.09 \\
    \addlinespace
ResNet18
        & 57.29\% & 0.947 & 0.951 & 0.950 & 0.951 & 64.36 & 74.17 & 50.82 & 28.37 \\
    \addlinespace
DenseNet161
        & 65.13\% & 0.941 & 0.950 & 0.950 & 0.950 & 56.61 & 79.09 & 26.74 & 16.81 \\
    \addlinespace
VGG16
        & 58.81\% & 0.934 & 0.949 & 0.949 & 0.951 & 66.23 & 67.14 & 41.49 & 24.12 \\
    \addlinespace
ShuffleNet
        & 56.04\% & 0.937 & 0.951 & 0.949 & 0.950 & 136.52 & 143.00 & 92.51 & 47.89 \\
    \addlinespace
Inception
        & 57.59\%  & 0.939 & 0.950 & 0.951 & 0.951 & 240.92 & 283.44 & 68.85 & 44.05 \\
    % \addlinespace

    \bottomrule
\end{tabular}  }
\label{eff_v2_append}
\end{table}

\section{Adaptivity in Least Ambiguous Set-valued Classification}
\label{lac_comp_append}
Least Ambiguous Set-valued (LAS) classification method~\citep{cp_lac} uses one minus the temperature-scaled softmax scores as its non-conformity function to find the optimal threshold and include the labels in the sets accordingly. In~\tableref{sat_result_append}, we report SSCV as the adaptivity metric and SAT as the quality metric for both ECP and LAS along with their average marginal coverage and set sizes. We observe that both ECP and LAS produce small and highly efficient sets although having negligible differences in different models that are caused by stochasticity and finite validation dataset. However, unlike ECP, LAS sacrifices the set adaptivity by producing the smallest possible set sizes. In terms of adaptiveness, ECP shows significantly lower SSCV compared to LAS that demonstrates lower deviations from the exact conditional coverage in ECP (lower coverage violation). Therefore, ECP has higher SAT scores compared to LAS indicating a considerably better trade-off between set size and adaptivity (higher quality) in the practical predictions.  

\begin{table}[t]
\centering
\caption{The performance and quality of prediction sets in LAS compared to ECP on ImageNet-Val data in different methods and model architectures with $\delta=0.1$}
\scalebox{1.0}{
\begin{tabular}{l cccc cc>{\bfseries}c>{\bfseries}c}
    \toprule
\multirow{0}{*}{\textbf{Model}} 
        & \multicolumn{4}{c}{\textbf{LAS}} & \multicolumn{4}{c}{\textbf{ECP (Ours)}} \\
    \cmidrule(lr){2-5} \cmidrule(lr){6-9}
        & Coverage & Size & SSCV & SAT & Coverage & Size & SSCV & SAT \\
    \midrule
ResNeXT101 
        & 0.900 & 1.65 & 0.150 & 0.511 & 0.900 & 1.65 & 0.072 & 0.560 \\
    \addlinespace
ResNet152
        & 0.901 & 1.75 & 0.141 & 0.490 & 0.900 & 1.76 & 0.077 & 0.520 \\
    \addlinespace
ResNet101
        & 0.900 & 1.85 & 0.147 & 0.465 & 0.899 & 1.83 & 0.076 & 0.486 \\
    \addlinespace
ResNet50
        & 0.900 & 2.04 & 0.131 & 0.426 & 0.900 & 2.06 & 0.092 & 0.439 \\
    \addlinespace
ResNet18
        & 0.900 & 3.65 & 0.207 & 0.216 & 0.900 & 3.66 & 0.075 & 0.254 \\
    \addlinespace
DenseNet161
        & 0.901 & 1.85 & 0.138 & 0.464 & 0.901 & 1.89 & 0.070 & 0.489 \\
    \addlinespace
VGG16
        & 0.899 & 2.97 & 0.202 & 0.267 & 0.900 & 2.95 & 0.067 & 0.308 \\
    \addlinespace
ShuffleNet
        & 0.900 & 4.05 & 0.196 & 0.198 & 0.900 & 4.06 & 0.138 & 0.228 \\
    \addlinespace
Inception
        & 0.900 & 3.96 & 0.185 & 0.205 & 0.900 & 4.05 & 0.132 & 0.242 \\
    % \addlinespace

    \bottomrule
\end{tabular}  }
\label{sat_result_append}
\end{table}

% \begin{table}[t]
% \centering
% \caption{The average marginal coverage and set size of images stratified by their difficulty levels in LAC compared to ECP on ImageNet-Val using ResNet-152 with $\delta=0.1$}
% \scalebox{0.75}{
% \begin{tabular}{l ccc ccc}
%     \toprule
% \multirow{0}{*}{\textbf{Difficulty}} & \multicolumn{3}{c}{\textbf{ECP (Ours)}} 
%             & \multicolumn{3}{c}{\textbf{LAC}} \\
%     \cmidrule(lr){2-4} \cmidrule(lr){5-7} 
%         & Count & Coverage & Size & Count & Coverage & Size \\
%     \midrule
% 1
%         & 27336 & 0.999 & 1.47 & 27392 & 0.999 & 1.50 \\
%     \addlinespace
% 2 to 3
%         & 4573 & 0.817 & 2.56 & 4591 & 0.842 & 2.70 \\
%     \addlinespace
% 4 to 6
%         & 1239 & 0.203 & 2.84 & 1200 & 0.280 & 3.18 \\
%     \addlinespace
% 7 to 10
%         & 621 & 0.005 & 2.64 & 590 &  0.006 & 3.10 \\
%     \addlinespace
% 11 to 100
%         & 1092 & 0 & 2.32 & 1095 &  0 & 2.87 \\
%     \addlinespace
% 101 to 1000
%         & 139 & 0 & 2.00 & 132 &  0 & 2.56 \\
%     % \addlinespace

%     \bottomrule
% \end{tabular}  }
% \label{diff_result_append}
% \end{table}

\end{document}